\documentclass[runningheads]{llncs}

\usepackage{eccv}

\usepackage{eccvabbrv}

\usepackage{graphicx}
\usepackage{booktabs}

\usepackage[accsupp]{axessibility}  

\usepackage{hyperref}

\usepackage{orcidlink}

\usepackage{xcolor}
\usepackage[table]{xcolor} 
\usepackage{multirow}

\usepackage{tablefootnote}
\usepackage{pifont}     
\usepackage{makecell}

\usepackage[export]{adjustbox}
\usepackage{tabularx}      
\usepackage{ragged2e} 
\usepackage{seqsplit}

\begin{document}

\title{Zoom-IQA: Image Quality Assessment with Reliable Region-Aware Reasoning}

\titlerunning{Zoom-IQA}

\author{Guoqiang Liang\inst{1}\orcidlink{0000-0003-4790-7075} \and
Jianyi Wang\inst{1}\orcidlink{0000-0001-7025-3626} \and
Zhonghua Wu\inst{2}\orcidlink{0000-0001-8688-9098} \and Shangchen Zhou\inst{1}$^{*}$\orcidlink{0000-0001-8201-8877} \and Chen Change Loy\inst{1}$^{*}$\orcidlink{0000-0001-5345-1591} }

\authorrunning{G. Liang et al.}

\institute{S-Lab, Nanyang Technological University, Singapore \\
\and SenseTime Research, Singapore \\
\small{Project Page: \url{https://ethanliang99.github.io/ZOOMIQA-Projectpage/}}}

\maketitle

{
  \renewcommand{\thefootnote}%
    {\fnsymbol{footnote}}
  \footnotetext[1]{Corresponding author}
}

\begin{abstract}
Image Quality Assessment (IQA) is a long-standing problem in computer vision. 
Previous methods typically focus on predicting numerical scores without explanation or provide low-level descriptions lacking precise scores.
Recent reasoning-based vision language models (VLMs) have shown strong potential for IQA, enabling joint generation of quality descriptions and scores.
However, we notice that existing VLM-based IQA methods tend to exhibit unreliable reasoning due to their limited capability of integrating visual and textual cues.
In this work, we introduce Zoom-IQA, a VLM-based IQA model to explicitly emulate key cognitive behaviors: uncertainty awareness, region reasoning, and iterative refinement.
Specifically, we present a two-stage training pipeline: 
1) supervised fine-tuning (SFT) on our Grounded-Rationale-IQA (GR-IQA) dataset to teach the model to ground its assessments in key regions; 
and 2) reinforcement learning (RL) for dynamic policy exploration, primarily stabilized by our KL-Coverage regularizer to prevent reasoning and scoring diversity collapse, and supported by a Progressive Re-sampling Strategy to mitigate annotation bias.
Extensive experiments show that Zoom-IQA achieves improved robustness, explainability, and generalization. 
The application to downstream tasks, such as image restoration, further demonstrates the effectiveness of Zoom-IQA.
\end{abstract}

\section{Introduction}
Image Quality Assessment (IQA) is a fundamental task in computer vision, aiming to evaluate the perceptual quality of images in alignment with human perception.
Its importance has grown rapidly as IQA models increasingly serve as critical perceptual reward signals for improving modern algorithms.
Specifically, IQA serves as a component in frameworks like Reinforcement Learning (RL) from Human Feedback (RLHF)~\cite{rombach2022high,dhariwal2021diffusion,wang2025unified,he2024videoscore} to align outputs with human preferences.
Similarly, in image restoration, IQA scores are used as differentiable rewards~\cite{sun2024rfsr} or in Direct Preference Optimization (DPO)~\cite{wu2025dp,cai2025dspo} to guide models toward perceptually superior results.

\begin{figure}[!t]
    \centering
    \includegraphics[width=1\linewidth]{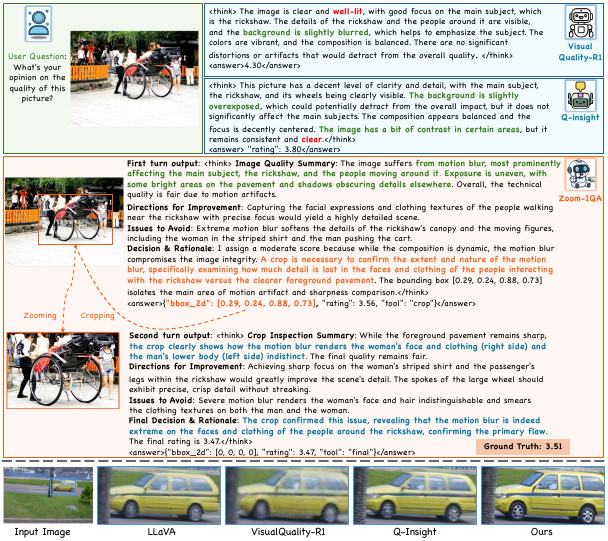}
    \vspace{-3mm}
    \caption{\textbf{(Upper)} Current IQA methods are \textbf{non-interactive}, leading to inferior assessments. They either spot only partial flaws (e.g., \textcolor[HTML]{3B7D23}{\textbf{slightly overexposed}} or \textcolor[HTML]{3B7D23}{\textbf{slightly blurred}}) or make factually incorrect claims (\textcolor{red}{\textbf{clear}} and \textcolor{red}{\textbf{well-lit}}), resulting in erroneous judgments. Our Zoom-IQA first hypothesizes flaws (\textcolor[HTML]{3B7D23}{green text}), then grounds them by cropping (\textcolor[HTML]{E97132}{orange text}), and finally verifies the degradation (\textcolor[HTML]{0B76A0}{blue text}). The predicted score rankings for this image are top 37.1\% for Q-Insight, top 35.1\% for VisualQuality-R1, top 34.9\% for ours, and top 31.9\% for the ground truth. \textbf{(Lower)} Our model's reasoning outputs also benefit downstream tasks, such as text-guided image restoration with SUPIR~\cite{yu2024scaling}. Our prompt enables a far superior restoration compared to those guided by other IQA methods or SUPIR's default VLM, LLaVA-1.5-13b~\cite{liu2023visual}.}
    \label{fig:teaser}
\vspace{-10pt}
\end{figure}

The emergence of vision language models (VLMs) like CLIP~\cite{radford2021learning} opened a promising direction for IQA~\cite{wang2022exploring}, further advanced by large-scale VLMs~\cite{touvron2023llama,liu2023visual,bai2025qwen2} that leverage broad knowledge to better align with human perception.
Existing VLM-based IQA methods fall into two categories:
(1) \textit{Score-based methods} (\eg, Q-Align~\cite{wu2023q} and DeQA-Score~\cite{you2025teaching}), which emphasize accurate scores but lack the ability to provide textual descriptions; and 
(2) \textit{Description-based methods} (\eg, DepictQA series~\cite{you2024depicting, you2024descriptive}), which offer detailed explanations but rely on SFT data with descriptions generated from ground-truth synthetic distortions.
To bridge this gap, recent methods like Q-Insight~\cite{li2025q} and VisualQuality-R1~\cite{wu2025visualquality} introduce RL to unify quality scoring and textual reasoning using only score labels as rewards.

Despite impressive capabilities, these VLM-based IQA methods remain \textit{non-interactive}. 
They generate responses in a single pass without any mechanism for iterative visual refinement or correction. 
As highlighted in complex visual tasks (\eg, object detection~\cite{liu2025visual,shen2025vlm} and VQA~\cite{yu2025perception,meng2025mm}), the lack of intermediate visual grounding may lead to unreliable responses and reasoning, especially under complex scenarios.
Similarly, the nature of IQA requires subjects to interpret complex images by ``zooming in'' on key regions. 
The importance of this behavior is highlighted by the DiffIQA dataset~\cite{chen2025toward}, which provided annotators with a zoom-in feature to inspect details.
Such a region-aware interaction plays a helpful role in understanding the quality of an image, but is absent in existing VLM-based IQA models, confining their reasoning to the text domain and limiting effective use of visual information, as shown in Fig.~\ref{fig:teaser}.

Enabling a VLM-based IQA method to dynamically ``crop and zoom'' for iterative region-aware assessment faces two core challenges. 
(1)~\textit{Region-aware Learning.} The model must learn where to focus and how to transform regions (\eg, crop, zoom) based on its own partial textual deliberations, similar to the grounding in VQA~\cite{zheng2025deepeyes, shao2024visual}. 
However, grounding IQA is more challenging than VQA. In VQA, grounding is often explicit, \ie, an answer can be tied to discrete, localizable semantic objects, for which extensive annotations are available~\cite{zheng2025deepeyes, shao2024visual}. In contrast, an IQA score is a holistic judgment aggregated from numerous, complex factors. The core difficulty is the lack of supervision specifying which regions a human prioritized to arrive at their final score. While recent IQA grounding datasets~\cite{chen2024q, chen2024grounding} provide static distortion masks, these methods fail to reveal the criticality of those regions to the overall human assessment or capture the dynamic reasoning path that led to the final judgment.
(2)~\textit{Self-Guided Reasoning Policy.} 
The model must learn a dynamic policy on when to trigger detailed visual inspection rather than relying on random exploration or preset rules for zooming. 
This requires an unsupervised iterative cognitive process, \ie, performing a holistic assessment, identifying its own uncertainty about a specific region, and only then deciding to ``zoom in'' for refinement.

To bridge such gaps, we make \textbf{two primary contributions}.
\textbf{First}, we introduce Grounded-Rationale-IQA (GR-IQA), a fine-grained dataset curated to facilitate the development of interleaved text-image Chain-of-Thought (CoT) reasoning.
Directly harnessing advanced VLMs such as Gemini~\cite{comanici2025gemini} to label IQA data with reasoning and scores can suffer from misalignment between visual inputs and reasoning outputs due to hallucination~\cite{leng2024mitigating,liu2024paying}.
Our GR-IQA is designed to avoid such hallucination by providing rationales that are verifiably grounded in visual regions.
Specifically, our curation pipeline introduces two key modules: (1) \textit{Visual Reliance Filtering (VRF)}, which enforces grounding by measuring the generative output shift (with vs. without the image); and (2) \textit{Hint-Augmented Consistency Filtering (HACF)}, which performs sample-level checks to filter hallucination-like descriptions.
\textbf{Second}, we propose Zoom-IQA (Zoomable Region Reasoning for Reliable Image Quality Assessment), a novel framework designed to enhance the reasoning reliability of IQA with region awareness.
Zoom-IQA is trained in two stages: it first learns formatted grounding (how to ``zoom'') via supervised fine-tuning (SFT) on our GR-IQA dataset, and then learns a dynamic policy (when to ``zoom'') via reinforcement learning (RL).
Such a learned policy allows Zoom-IQA to operate iteratively, moving beyond ``single-pass'' methods to identify uncertainty and refine its assessment, achieving truly interactive visual reasoning.
To stabilize the training process, we further propose the KL-Coverage regularizer, designed to prevent a collapse in reasoning path diversity, which often leads to a severe ``mode collapse''. 
A Progressive Re-sampling Strategy is also developed to mitigate bias from imbalanced annotations.

Our Zoom-IQA is evaluated across diverse datasets and IQA tasks, demonstrating superior performance over both conventional IQA metrics and recent SFT-driven large language models.
Moreover, Zoom-IQA exhibits impressive zero-shot generalization, such as effectively guiding image restoration models at test time, which highlights the robustness and real-world applicability of its region reasoning.

\section{Related Work}
\noindent {\bf Image Quality Assessment.}
Previous IQA works are broadly divided into full-reference (FR) and no-reference (NR) approaches, based on the availability of a pristine reference image. As our work does not require a reference, we focus on the more challenging NR-IQA task.
Conventional NR-IQA methods~\cite{ma2017learning,mittal2012no,mittal2012making,moorthy2010two,moorthy2011blind} relied on hand-crafted, degradation-aware features to predict the final quality score. Subsequent deep learning-based models~\cite{kang2014convolutional,bosse2017deep,su2020blindly,ke2021musiq,cheon2021perceptual,sun2022graphiqa,liu2017rankiqa,pan2018blind,zhu2020metaiqa} replaced this pipeline, directly predicting quality scores using end-to-end trainable neural networks.
\textit{Nonetheless, these models often suffer from significant performance degradation on out-of-distribution (OOD) data, limiting their practical usage and generalizability.}

\noindent {\bf Vision Language Models in Image Quality Assessment.}
Vision language models (VLMs)~\cite{radford2021learning,touvron2023llama,liu2023visual,bai2025qwen2} have been extensively studied in image quality assessment (IQA), leveraging their powerful cross-modal understanding and strong generalization capabilities. These works typically focus on one of two objectives: providing numerical quality scores~\cite{wang2022exploring,wu2023q,you2025teaching,zhu2024adaptive,liu2024dog} or generating visual quality descriptions~\cite{wu2024q,you2024depicting,you2024descriptive,chen2024grounding}.
Specifically, CLIP-IQA proposes to harness CLIP~\cite{radford2021learning} for image quality assessment from multiple aspects. DOG-IQA~\cite{liu2024dog} attempts to mimic the human evaluation process (\eg, zooming in to evaluate specific areas). However, these models lack the necessary reasoning capabilities and cannot dynamically decide which regions to inspect. A common way to alleviate such limitations is via pre-processing, \ie, using a pre-trained segmentation model to crop sub-images and then computing the final score as a weighted average of these crops.
Recent works like Q-Insight~\cite{li2025q}, VisualQuality-R1~\cite{wu2025visualquality}, and Q-Ponder~\cite{cai2025q} propose to employ reinforcement learning (RL) to leverage the reasoning capabilities of VLMs, enabling image quality rating as well as textual justifications.
Building upon this paradigm, subsequent advancements have expanded RL-based reasoning to video quality assessment~\cite{zhang2026vq} and employed contrastive learning to directly align images with the generalizable text representations learned via RL~\cite{zhao2025reasoning}.
However, the reasoning chains generated by these methods remain purely textual. They lack dynamic interaction with the image (\eg, cropping or zooming) to evaluate specific regions—a process crucial to human assessment. \textit{Consequently, visual evidence is insufficiently explored, and the models fail to ground their textual reasoning in verifiable image regions, limiting both the reliability and interpretability of their outputs.}

\noindent {\bf Vision Language Models with Multimodal Reasoning.}
Recent advances in enhancing the reasoning capabilities of VLMs have significantly improved their performance on challenging tasks, such as mathematical problem solving~\cite{meng2025mm,huang2025vision,zhang2025r1}, general VQA~\cite{yu2025perception,meng2025mm}, and object detection~\cite{liu2025visual,shen2025vlm,yu2025perception}.
However, these models typically generate reasoning chains composed solely of natural language. This text-only reasoning can be opaque and often lacks sufficient grounding in the visual input's fine-grained details.
To address this, recent works in general VQA~\cite{zheng2025deepeyes,zhang2025chain,su2025pixel,jiang2025vlm} propose to integrate evidence regions into the reasoning process. 
By equipping VLMs with capabilities like iterative zoom-in and region-of-interest selection, these methods demonstrated boosted performance.
Improved interpretability is further gained via visual-linguistic interactive reasoning.
However, these VQA methods generally rely on heavily annotated evidence regions (\eg, bounding boxes) for training. 
\textit{Such fine-grained regional labels and corresponding reasoning trajectories are expensive and lacking in the IQA domain.}

\begin{figure*}[t]
\centering
    \includegraphics[width=1\linewidth]{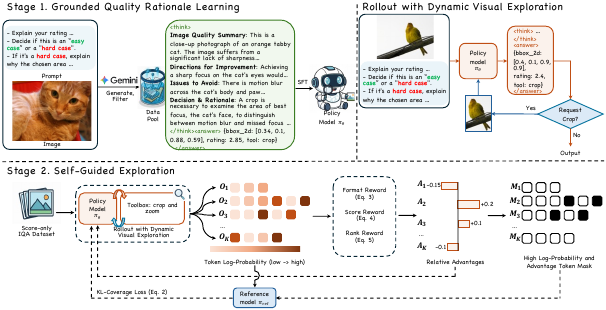}
    \vspace{-5pt}
    \caption{\textbf{An overview of our two-stage framework.} Stage \textbf{(1)}, \textbf{Grounded Quality Rationale Learning} (Sec.~\ref{Sec:cold_start}), first uses SFT to teach the model how to correctly execute the crop action. Stage \textbf{(2)}, \textbf{Self-Guided Exploration} (Sec.~\ref{Sec:grpo}), then uses RL to let the model learn what to crop, allowing it to discover regions that lead to a deeper understanding of image quality.}
    \label{fig:framework}
\centering
\vspace{-10pt}
\end{figure*}

\section{Methodology}
Our method, Zoom-IQA, is trained via a two-stage pipeline (illustrated in Fig.~\ref{fig:framework}):
\textbf{1)} Supervised Fine-Tuning for Grounded Quality Rationale Learning: We first leverage our GR-IQA dataset to teach the VLM the foundational ``how-to'' skills: grounding textual rationales in visual regions and executing the ``zoom'' action~(Sec.~\ref{Sec:cold_start}).
\textbf{2)} Reinforcement Learning for Self-Guided Exploration: We then use the RL-based training process (Group Relative Policy Optimization) to learn a dynamic policy that decides when to deploy these skills for iterative refinement~(Sec.~\ref{Sec:grpo}).

\subsection{Grounded Quality Rationale Learning}

\label{Sec:cold_start}
Recalling the challenges from the Introduction, grounding IQA is uniquely challenging. While VQA supervision can link answers to semantic regions, IQA supervision is often limited to static distortion masks. This static approach fails to capture the dynamic reasoning path of how or why specific regions influence the final holistic assessment.

\begin{figure*}[t]
\centering
    \includegraphics[width=0.99\linewidth]{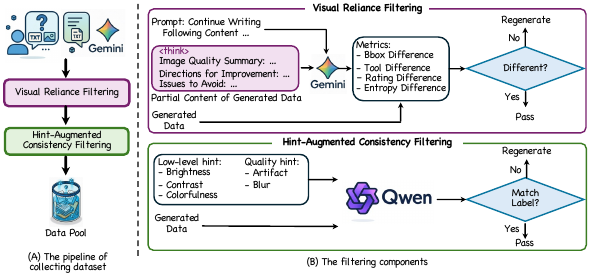}
    \caption{The GR-IQA dataset curation pipeline. It uses \textbf{(1)} Visual Reliance Filtering (VRF) to ensure visual grounding by measuring the output shift with vs.\ without the image, and \textbf{(2)} Hint-Augmented Consistency Filtering (HACF) to perform sample-level, hint-based checks for unfaithful text.}
    \label{fig:dataset}
\centering
\vspace{-10pt}
\end{figure*}
\subsubsection{Grounded-Rationale-IQA (GR-IQA) Dataset.}
To bridge this gap, we curate the Grounded-Rationale-IQA (GR-IQA) dataset with approximately 7,000 reasoning trajectories. This curation is performed through a novel pipeline (Fig.~\ref{fig:dataset}) that features our two key modules: \textbf{1)} Visual Reliance Filtering (VRF); \textbf{2)} Hint-Augmented Consistency Filtering (HACF).

\noindent {\bf Data Generation.}
We prompt the closed-source VLM, Gemini-2.5-pro~\cite{comanici2025gemini}, on KonIQ dataset images using a structured prompt. This compels the VLM to generate a two-part response: a textual rationale (within \texttt{<think>}) and a JSON action (within \texttt{<answer>}). The textual rationale is strictly constrained to a four-part format: (1) a holistic \textit{Image Quality Summary}; (2) \textit{Directions for Improvement}; (3) \textit{Issues to Avoid}; and (4) a \textit{Decision \& Rationale} where the model must decide if the image is an ``easy case'' (requiring a ``final'' tool) or a ``hard case'' (requiring a ``crop'' tool) and justify its choice. The \texttt{<answer>} block then contains the chosen ``tool'' (``final'' or ``crop''), the rating, and a conditional ``bbox''. This structured process forces the VLM to link its score to \textbf{regional evidence} and to \textbf{perform self-assessment on its own uncertainty} (\ie, ``zoom'' or not). This raw output forms the data for our GR-IQA dataset, which is then passed to our filtering modules (VRF and HACF).

\noindent \textbf{Visual Reliance Filtering (VRF).}
Despite strong language priors, VLMs can over-rely on textual co-occurrence and generate responses that are weakly grounded in the image~\cite{liu2024paying,leng2024mitigating}.
Detecting such hallucinations is non-trivial in our setting: (i) for closed-source VLMs, token-level log-probabilities are often unavailable, making probability-based contrastive analysis inapplicable; and (ii) online consistency checks that inject image perturbations are ill-suited for IQA, since added distortions can directly alter the target degradation to be assessed.
We therefore propose \emph{Visual Reliance Filtering} (VRF), which removes samples with low \emph{visual} reliance without modifying the input image.
Specifically, we contrast the VLM outputs under two inference conditions:
\textbf{(1)} conditioned on both the image $I$ and a partial rationale $R_A$, and
\textbf{(2)} conditioned only on $R_A$ (without $I$).
We measure their discrepancy using task-relevant signals (e.g., predicted quality score difference, localization overlap, and output uncertainty);
samples with overly small discrepancies are discarded, indicating that the response can be reproduced from text alone.

\noindent \textbf{Hint-Augmented Consistency Filtering (HACF).}
While VRF encourages visual dependence for the final \texttt{<answer>}, the intermediate rationale $R$ (the \texttt{<think>} block) may still contain statements unsupported by the visual evidence.
We thus employ a strong LLM rater, Qwen-2.5-32b~\cite{qwen2.5}, denoted as $\text{LLM}_{\text{Rater}}$, to perform a sample-level assessment of the rationale's image-consistency.
To aid this judgment, $\text{LLM}_{\text{Rater}}$ is provided with $(R, I)$ together with \emph{dataset-provided} low-level, non-semantic hints $H$ (e.g., brightness, sharpness, and color statistics).
The rater outputs a binary accept/reject decision:
$
D_{\text{HACF}} = \text{LLM}_{\text{Rater}}(R, I, H).
$
We only retain samples with $D_{\text{HACF}}=\text{``Pass''}$, ensuring that the training data exhibits image-consistent reasoning.
\textit{Additional examples of filtered-out samples are provided in the supplementary material.}

\subsubsection{Grounded Rationale Fine-Tuning.}
\label{Sec:sft_training}
Following the curation of our high-fidelity GR-IQA dataset, we proceed to the SFT stage. 
We fine-tune the VLM, parameterized by $\theta$, to auto-regressively generate the complete ground-truth response $S_{\text{structured}} = (R, A)$, given the image $I$ and the prompt $T_{\text{CoT}}$. 
This is achieved by minimizing the standard cross-entropy (CE) loss $\mathcal{L}_{\text{SFT}}$:
$
\mathcal{L}_{\text{SFT}}(\theta) = - \sum_{t=1}^{|S|} \log p_\theta(S_t \mid S_{<t}, I, T_{\text{CoT}})
$
where $S_t$ is the $t$-th token in the ground-truth sequence $S_{\text{structured}}$, and $|S|$ is the total length of the sequence.

\subsection{Self-Guided Exploration}
\label{Sec:grpo}

\noindent \textbf{KL-Coverage Regularizer.}
Recent studies on applying Reinforcement Learning (RL) to large language models (LLMs)~\cite{cui2025entropy,cheng2025reasoning} highlight a critical challenge: policy entropy often drops sharply at the onset of training, declining monotonically to near zero. This ``entropy collapse'' severely limits the model's ability to explore, leading to performance plateaus.
In the context of IQA, this issue is particularly detrimental. It leads to a collapse in the diversity of both reasoning paths and predicted rating scores. 
For instance, existing RL-based IQA methods, such as VisualQuality-R1~\cite{wu2025visualquality}, suffer from ``score collapse.'' On the KonIQ~\cite{hosu2020koniq} test set, this method's output unique score ratio is merely 2.04\%, in stark contrast to the 71.34\% of the ground-truth Mean Opinion Scores (MOS) distribution (when rounded to two decimal places).

To address this problem, we propose the KL-Coverage regularizer. 
This approach is inspired by the use of KL penalties to constrain policy updates~\cite{schulman2017proximal} and the recent finding that high covariance between action log-probabilities and logit changes leads to policy entropy collapse~\cite{cui2025entropy}. 
Our regularizer is thus designed to specifically suppress numerical tokens that exhibit this high covariance. 

Given a batch of $N$ rollout tokens, let $\pi_{\theta}(y_i | y_{<i})$ denote the policy's probability for token $y_i$ given its prefix $y_{<i}$, and let $A(y_i)$ be its associated advantage. We first compute the batch-level mean log-probability $\overline{\log \pi}$ and mean advantage $\overline{A}$:
$
\overline{\log \pi} = \frac{1}{N} \sum_{j=1}^{N} \log \pi_{\theta}(y_j | y_{<j}), \quad \overline{A} = \frac{1}{N} \sum_{j=1}^{N} A(y_j).
$
We then define a token-wise covariance score $Cov(y_i)$ as the centered cross-product:
\begin{equation}
Cov(y_i) = \left( \log \pi_{\theta}(y_i | y_{<i}) - \overline{\log \pi} \right) \left( A(y_i) - \overline{A} \right).
\end{equation}
\textit{Crucially, our regularizer mainly targets the numerical tokens responsible for the final score.}
We first define a candidate set $\mathcal{N}_{ans}$ comprising all numerical tokens within the \texttt{<answer>...</answer>} tags. 
We then rank the tokens in this candidate set $\mathcal{N}_{ans}$ by their $Cov(y_i)$ scores.
We define a binary mask $M_i$ for tokens $y_i \in \mathcal{N}_{ans}$, where $M_i=1$ if the token is in the top-$p$ proportion (e.g., $p=0.02$), and $M_i=0$ otherwise. Here, $p$ is a hyperparameter defining the fraction of these candidate tokens to be regularized.

Finally, we impose the KL penalty only on these selected tokens (where $M_i=1$). The KL-Coverage loss, $\mathcal{L}_{KLC}$, is computed as the mean KL divergence between the old policy $\pi_{\theta_{\text{old}}}$ and the current policy $\pi_{\theta}$, averaged only over these masked-in tokens:
\begin{equation}
\mathcal{L}_{KLC} = \frac{\sum_{y_i \in \mathcal{N}_{ans}} M_i \cdot D_{KL}\left(\pi_{\theta_{\text{old}}}(y_i | y_{<i}) \middle\| \pi_{\theta}(y_i | y_{<i})\right)}{\sum_{y_i \in \mathcal{N}_{ans}} M_i}.
\end{equation}

\noindent \textbf{Progressive Re-sampling Strategy.}
Our training data suffers from a long-tailed score distribution, leading to poor performance on scarce score intervals (\eg, very high or low quality). 
To mitigate this data bias, we adopt a multi-stage re-sampling strategy. The model is first trained on the original data distribution, and in subsequent stages, we progressively increase the sampling frequency of these under-represented score intervals. 
This allows the model to first learn the general distribution and then fine-tune on rarer data, improving its robustness across the entire score range.

\noindent \textbf{Format Reward.} This reward verifies that the model's output strictly adheres to our required structured reasoning format. Specifically, the reasoning process, enclosed in \texttt{<think>...</think>} tags, must explicitly articulate key components such as \texttt{"Directions for Improvement"} and \texttt{"Issues to Avoid"}. Furthermore, the final decision must be in a structured \texttt{<answer>...</answer>} tag, containing elements like \texttt{bbox\_2d} and \texttt{rating}. If any of these formats are incorrect, the format score is 0. Only when all formats are correct can the model achieve the format score of 1.0 as defined:

\begin{equation}
    \begin{aligned}
    R_{\text{format}}(O) = \begin{cases} 
    1.0 & \text{if } O \text{ satisfies all format requirements} \\ 
    0 & \text{otherwise}
\end{cases}.
    \end{aligned}
    \label{eq:format_reward}
\end{equation}
\noindent \textbf{Score Reward.} This reward encourages the model to predict a quality rating $r_{\text{pred}}$ that is close to the ground-truth score $r_{\text{gt}}$. We define this as a continuous Gaussian reward based on their difference:

\begin{equation}
    \begin{aligned}
    R_{\text{score}} = \exp\left(-\frac{(r_{\text{pred}} - r_{\text{gt}})^2}{2\sigma^2}\right), \\
    \end{aligned}
    \label{eq:score_reward}
\end{equation}
where $\sigma$ is a hyperparameter controlling the sensitivity of the reward.

\noindent \textbf{Rank Reward.} To ensure the model learns relative quality ordering, we define a rank reward $R_{\text{rank}}(x_i)$ based on pairwise comparisons within a batch, inspired by the Thurstone model \cite{thurstone2017law}. It is computed as:

\begin{equation}
    \begin{aligned}
    R_{\text{rank}}(x_i) = \frac{1}{B-1} \sum_{j \neq i} \left( \sqrt{\hat{p}_{ij} p^*_{ij}} + \sqrt{(1-\hat{p}_{ij})(1-p^*_{ij})} \right),
    \end{aligned}
    \label{eq:rank_reward}
\end{equation}
where $p^*_{ij} = p(x_i, x_j)$ is the ground-truth preference derived from MOS, indicating whether $\text{MOS}(x_i) > \text{MOS}(x_j)$, and $\hat{p}_{ij}$ is the model's predicted preference probability, calculated using the Thurstone model:
$ \hat{p}_{ij} = \Phi\left(\frac{\mu_i - \mu_j}{\sqrt{v_i + v_j}}\right).$
Here, $\mu$ and $v$ represent the estimated mean and variance of the model's rating distribution for an input, and $\Phi$ is the standard normal CDF.

\noindent \textbf{Total Reward.}
The total reward $R_{\text{total}}$ for a trajectory comprises several components designed to guide the behavior of Zoom-IQA, formulated as:
\begin{equation}
    R_{\text{total}} = R_{\text{format}} + \alpha_{score} R_{\text{score}} + \alpha_{rank} R_{\text{rank}},
\end{equation}
where $\alpha_{score}$ and $\alpha_{rank}$ are coefficients that balance the importance of score prediction and rank consistency. In our experiments, we set $\alpha_{score} = 1$ and $\alpha_{rank} = 2$.

\begin{figure*}[!t]
\centering
    \includegraphics[width=1\linewidth]{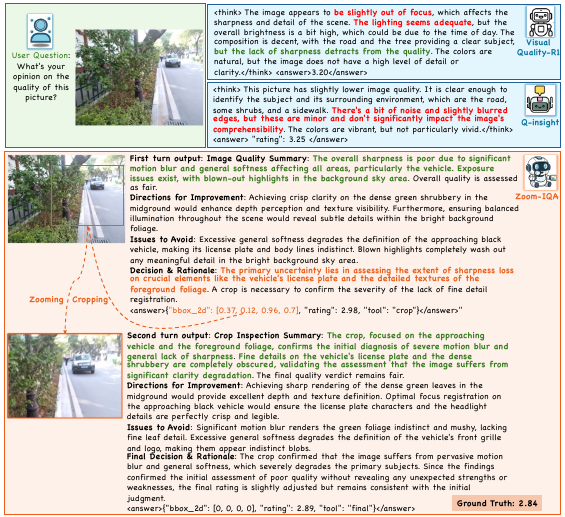}
    \vspace{-5mm}
    \caption{Qualitative comparison of Zoom-IQA with competing methods (Q-Insight~\cite{li2025q}, VisualQuality-R1~\cite{wu2025visualquality}). We highlight \textcolor[HTML]{3B7D23}{\textbf{correct}} descriptions, \textcolor{red}{\textbf{incorrect}} descriptions, and the \textcolor[HTML]{E97132}{\textbf{uncertainty-aware}} reasoning unique to our model. Specifically, the predicted score rankings for this image are top 42.5\% for Q-Insight, top 44.7\% for VisualQuality-R1, top 50.4\% for ours, and top 54.3\% for the ground truth.}
    \label{fig:qualitative_result}
\centering
\vspace{-3mm}
\end{figure*}

\begin{table*}[t]
\vspace{-2mm}
\centering
\caption{PLCC / SRCC comparison on the score regression tasks between our method and other competitive IQA methods. All methods except handcrafted ones are trained on the \textbf{KonIQ dataset}.}
\vspace{-1mm}
\label{tab:iqa_comparison}
\renewcommand{\arraystretch}{1.35}
\renewcommand{\tabcolsep}{1.45mm}

\resizebox{\textwidth}{!}{%
\begin{tabular}{@{}l|l|l|ccccccc@{}}
\toprule
\textbf{Metric} & \textbf{Category} & \textbf{Methods}
& \textbf{KonIQ} & \textbf{SPAQ} & \textbf{KADID} & \textbf{PIPAL} & \textbf{LiveW} & \textbf{AGIQA} & \textbf{CSIQ} \\
\midrule

\multirow{14}{*}{\textbf{PLCC}}
& \multirow{2}{*}{Handcrafted}
& NIQE~\cite{mittal2012making}                & 0.533 & 0.679 & 0.468 & 0.195 & 0.493 & 0.560 & 0.718 \\
& & BRISQUE~\cite{mittal2012no}               & 0.225 & 0.490 & 0.429 & 0.267 & 0.361 & 0.541 & 0.740 \\
\cmidrule(lr){2-10}

& \multirow{5}{*}{Non-VLM}
& NIMA~\cite{talebi2018nima}                  & 0.896 & 0.838 & 0.532 & 0.390 & 0.814 & 0.715 & 0.695 \\
& & HyperIQA~\cite{su2020blindly}             & 0.917 & 0.791 & 0.506 & 0.410 & 0.772 & 0.702 & 0.752 \\
& & DBCNN~\cite{zhang2018blind}               & 0.884 & 0.812 & 0.497 & 0.384 & 0.773 & 0.730 & 0.586 \\
& & MUSIQ~\cite{ke2021musiq}                  & 0.924 & 0.868 & 0.575 & 0.431 & 0.789 & 0.722 & 0.771 \\
& & ManIQA~\cite{yang2022maniqa}              & 0.849 & 0.768 & 0.499 & 0.457 & 0.849 & 0.723 & 0.623 \\
\cmidrule(lr){2-10}

& \multirow{7}{*}{\makecell[c]{VLM \\ (w/o \& w/ \\ reasoning)}}
& CLIP-IQA+~\cite{wang2022exploring}          & 0.909 & 0.866 & 0.653 & 0.427 & 0.832 & 0.736 & 0.772 \\
& & C2Score~\cite{zhu2024adaptive}            & 0.923 & 0.867 & 0.500 & 0.354 & 0.786 & 0.777 & 0.735 \\
& & Q-Align~\cite{wu2023q}                    & 0.941 & 0.886 & 0.674 & 0.403 & 0.853 & 0.772 & 0.671 \\
& & DeQA-Score~\cite{you2025teaching}               & 0.953 & 0.895 & 0.694 & 0.472 & 0.892 & 0.809 & 0.787 \\
\cline{3-10}
& & Q-Insight~\cite{li2025q}                  & 0.918 & 0.903 & 0.702 & 0.458 & 0.870 & 0.816 & 0.685 \\
& & VisualQuality-R1~\cite{wu2025visualquality}& 0.910 & 0.889 & 0.703 & 0.451 & 0.856 & 0.817 & 0.768 \\
& & Zoom-IQA (Ours)                           & 0.938 & 0.902 & 0.701 & 0.468 & 0.887 & 0.816 & 0.797 \\
\midrule

\multirow{14}{*}{\textbf{SRCC}}
& \multirow{2}{*}{Handcrafted}
& NIQE~\cite{mittal2012making}                & 0.530 & 0.664 & 0.405 & 0.161 & 0.449 & 0.533 & 0.628 \\
& & BRISQUE~\cite{mittal2012no}               & 0.226 & 0.406 & 0.356 & 0.232 & 0.313 & 0.497 & 0.556 \\
\cmidrule(lr){2-10}

& \multirow{5}{*}{Non-VLM}
& NIMA~\cite{talebi2018nima}                  & 0.859 & 0.856 & 0.535 & 0.399 & 0.771 & 0.654 & 0.649 \\
& & HyperIQA~\cite{su2020blindly}             & 0.906 & 0.788 & 0.468 & 0.403 & 0.749 & 0.640 & 0.717 \\
& & DBCNN~\cite{zhang2018blind}               & 0.875 & 0.806 & 0.484 & 0.381 & 0.755 & 0.641 & 0.572 \\
& & MUSIQ~\cite{ke2021musiq}                  & 0.929 & 0.863 & 0.556 & 0.431 & 0.830 & 0.630 & 0.710 \\
& & ManIQA~\cite{yang2022maniqa}              & 0.834 & 0.758 & 0.465 & 0.452 & 0.832 & 0.636 & 0.627 \\
\cmidrule(lr){2-10}

& \multirow{7}{*}{\makecell[c]{VLM \\ (w/o \& w/ \\ reasoning)}}
& CLIP-IQA+~\cite{wang2022exploring}          & 0.895 & 0.864 & 0.654 & 0.419 & 0.805 & 0.685 & 0.719 \\
& & C2Score~\cite{zhu2024adaptive}            & 0.910 & 0.860 & 0.453 & 0.342 & 0.772 & 0.671 & 0.705 \\
& & Q-Align~\cite{wu2023q}                    & 0.940 & 0.887 & 0.684 & 0.419 & 0.860 & {0.735} & 0.737 \\
& & DeQA-Score~\cite{you2025teaching}               & {0.941} & {0.896} & {0.687} & {0.478} & {0.879} & 0.729 & 0.744 \\
\cline{3-10}
& & Q-Insight~\cite{li2025q}                  & 0.895 & 0.903 & 0.702 & 0.435 & 0.839 & 0.766 & 0.640 \\
& & VisualQuality-R1~\cite{wu2025visualquality}& 0.896 & 0.892 & 0.712 & 0.441 & {0.827} & 0.760 & {0.707} \\
& & Zoom-IQA (Ours)                           & 0.922 & 0.900 & 0.700 & 0.465 & 0.870 & 0.765 & 0.754 \\
\bottomrule
\end{tabular}%
}
\vspace{-3mm}
\end{table*}

\begin{table*}[h]
\centering
\caption{Quantitative results for image quality description. We evaluate four metrics: Accuracy (Acc.), Reasonableness (Reason.), Completeness (Compl.), and Confidence (Conf.) using closed-source VLM evaluators: Gemini-2.5-Flash and GPT-5-mini.}
\renewcommand{\arraystretch}{1.15}
\renewcommand{\tabcolsep}{1.7mm}


\resizebox{\textwidth}{!}{
\begin{tabular}{llcccccccc}
\toprule
\multirow{2}{*}{Dataset} & \multirow{2}{*}{Method}
& \multicolumn{4}{c}{Gemini-2.5-flash} & \multicolumn{4}{c}{GPT-5-mini} \\
\cmidrule(lr){3-6} \cmidrule(lr){7-10}
& & Acc. $\uparrow$ & Reason. $\uparrow$ & Compl. $\uparrow$ & Conf. $\uparrow$
  & Acc. $\uparrow$ & Reason. $\uparrow$ & Compl. $\uparrow$ & Conf. $\uparrow$ \\
\midrule
\multirow{4}{*}{KonIQ}
& DepictQA            & 5.40 & 5.49 & 5.51 & 7.96 & 4.54 & 5.09 & 4.41 & 7.80 \\
& VisualQuality-R1  & 7.29 & 7.60 & 7.29 & 7.57 & 6.10 & 6.05 & 6.02 & 6.79 \\
& Q-Insight                      & 7.17 & 7.44 & 7.08 & 6.93 & 5.32 & 5.74 & 5.29 & 6.15 \\
& Zoom-IQA (Ours)                               & 8.72 & 8.80 & 8.30 & 8.60 & 6.93 & 6.93 & 6.61 & 7.98 \\
\midrule
\multirow{4}{*}{SPAQ}
& DepictQA             & 6.04 & 6.39 & 6.14 & 7.86 & 4.80 & 5.08 & 4.46 & 6.87 \\
& VisualQuality-R1  & 8.32 & 8.35 & 7.70 & 7.55 & 6.61 & 6.67 & 5.51 & 6.82 \\
& Q-Insight                      & 7.84 & 8.02 & 7.51 & 6.98 & 6.22 & 6.35 & 5.54 & 5.90 \\
& Zoom-IQA (Ours)                               & 8.63 & 8.69 & 8.47 & 8.63 & 6.97 & 7.27 & 6.79 & 7.99 \\
\bottomrule
\end{tabular}
}
\label{tab:reasoning_description}
\end{table*}

\section{Experiments}
\subsection{Experimental Settings}

\noindent \textbf{Implementation Details.}
We initialize Qwen2.5-VL-7b~\cite{bai2025qwen2} as our base model during the first cold start stage, in which training is performed with a batch size of 2, 4 gradient accumulation steps, a learning rate of $2.5 \times 10^{-6}$, and a warm-up ratio of 0.3. For GRPO, we train the finetuned model after the cold start stage with a batch size of 1, 2 gradient accumulation steps, a learning rate of $1 \times 10^{-6}$, and a KL penalty coefficient of $\beta = 0.04$. The number of generated responses N is set to 8. 

\noindent \textbf{Datasets and Metrics.}
For the first cold start stage, we applied SFT with our collected high-quality CoT datasets using cross-entropy loss.
For the score regression task, we conduct training and evaluation on seven IQA datasets grouped into three categories: (1) In-the-wild datasets, including KonIQ~\cite{hosu2020koniq}, SPAQ~\cite{fang2020perceptual}, and LIVE-Wild~\cite{ghadiyaram2015live}; (2) Synthetic distortion datasets, including KADID~\cite{lin2019kadid}, PIPAL~\cite{jinjin2020pipal}, and CSIQ~\cite{larson2010most}; (3) AI-generated image datasets, including AGIQA~\cite{li2023agiqa}.
We adopt the Pearson linear correlation coefficient (PLCC) and Spearman rank-order correlation coefficient (SRCC) as metrics to evaluate performance on the score regression task, following previous works~\cite{you2025teaching,li2025q}.

\subsection{Comparison and Evaluation}
\noindent \textbf{Image Quality Score Regression.}
We compare our method with SOTA IQA methods in three different categories: \textbf{(I)} handcrafted, including NIQE~\cite{mittal2012making} and BRISQUE~\cite{mittal2012no}; 
\textbf{(II)} deep learning-based, NIMA~\cite{talebi2018nima}, HyperIQA~\cite{su2020blindly}, DBCNN~\cite{zhang2018blind}, MUSIQ~\cite{ke2021musiq}, and ManIQA~\cite{yang2022maniqa}; 
\textbf{(III)} VLM-based models, CLIP-IQA+~\cite{wang2022exploring}, C2Score~\cite{zhu2024adaptive}, Q-Align~\cite{wu2023q}, DeQA-Score~\cite{you2025teaching}, VisualQuality-R1~\cite{wu2025visualquality}, and Q-Insight~\cite{li2025q}. Since VisualQuality-R1~\cite{wu2025visualquality} did not report a KonIQ-only trained model, we retrained it with its official training code.
As shown in Table~\ref{tab:iqa_comparison}, our approach achieves comparable performance to existing baselines across various synthetic and real-world benchmarks. When comparing with state-of-the-art IQA methods~\cite{li2025q,wu2025visualquality} w/ reasoning capability, our Zoom-IQA presents consistently superior performance across almost all the benchmarks.
Furthermore, a qualitative comparison demonstrates the superiority of our region-aware reasoning over competing methods (Fig.~\ref{fig:qualitative_result}).

\noindent \textbf{Image Quality Reasoning.} 
To validate the effectiveness and accuracy of our reasoning chains, we follow common practices~\cite{chen2024mllm,jiang2025mme} to employ a VLM-as-judge evaluation methodology on the KonIQ and SPAQ datasets. We utilize two powerful, closed-source VLMs (Gemini-2.5-Flash~\cite{comanici2025gemini} and GPT-5-mini~\cite{singh2025openai}) as evaluators, which are tasked to score the generated descriptions on a 1-to-9 scale across four key criteria: Accuracy, Reasonableness, Completeness, and Confidence. To ground the assessments and ensure objectivity, the VLMs are prompted with the image, the generated reasoning chain, and corresponding low-level image indicators (e.g., brightness, sharpness) for cross-referencing. The detailed definitions of each metric and the full prompt structure are provided in the Appendix.
As shown in Table \ref{tab:reasoning_description}, our method (Zoom-IQA) consistently and significantly outperforms all baselines \cite{you2024depicting, wu2025visualquality, li2025q} across both datasets and under the scrutiny of both VLM evaluators, indicating the superiority of our reasoning reliability.

\begin{figure*}[t]
\centering
    \includegraphics[width=1.0\linewidth]{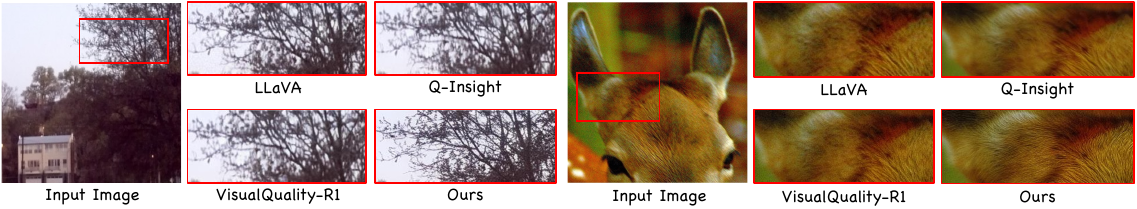}
    \vspace{-5mm}
    \caption{Qualitative evaluation of reasoning quality on the image restoration task.}
    \vspace{-6mm}
    \label{fig:restoration}
\centering
\end{figure*}

\noindent \textbf{Reasoning-guided Restoration.}
High-quality visual reasoning provides reliable guidance for downstream tasks such as generative image restoration~\cite{wang2024exploiting,yu2024scaling,wang2025seedvr,wang2025seedvr2}.
To validate the efficacy and transferability of Zoom-IQA's reasoning capability, we leverage its outputs to guide two state-of-the-art text-guided restoration frameworks: SUPIR~\cite{yu2024scaling} and DreamClear~\cite{ai2024dreamclear}.
Both frameworks follow a cascaded paradigm where initial results from a lightweight restoration network are conditioned by VLM-generated guidance.
For the SUPIR framework, we replace its default prompt generator (LLaVA-1.5-13b~\cite{liu2023visual}) with the textual reasoning outputs derived from Q-Insight~\cite{li2025q}, VisualQuality-R1~\cite{wu2025visualquality}, and Zoom-IQA. As illustrated in Fig.~\ref{fig:restoration}, Zoom-IQA generates highly specific and context-aware guidance, enabling SUPIR to accurately restore fine-grained, detailed textures that are otherwise overlooked or poorly reconstructed by competing methods.
Concurrently, we extend our evaluation to DreamClear~\cite{ai2024dreamclear}. 
This choice is motivated by its T5 text encoder~\cite{raffel2020exploring}, which supports a longer context window than the CLIP encoder~\cite{radford2021learning} used in SUPIR, thereby accommodating the detailed reasoning prompts generated by Zoom-IQA.
Under this framework, we compare the default captions from LLaVA-1.6-13b~\cite{liu2023visual} against the reasoning content from VisualQuality-R1, Q-Insight, and Zoom-IQA's reasoning trajectory.

To quantitatively evaluate the quality of the reasoning guidance across these different configurations, we conduct comprehensive experiments using SUPIR on DIV2K (synthetic) and DreamClear on RealLQ250 (real-world), as summarized in Tab.~\ref{tab:restoration_table}.
\begin{table}[t]
\centering
\caption{Quantitative results of image restoration using different reasoning guidance.}
\vspace{-2mm}
\label{tab:restoration_table}
\renewcommand{\arraystretch}{1.15}
\setlength{\tabcolsep}{1.7mm}
\resizebox{\linewidth}{!}{ 
\begin{tabular}{lccccccc}
\toprule
\multirow{2}{*}{Guidance} 
& \multicolumn{3}{c}{DIV2K (SUPIR)} 
& \multicolumn{4}{c}{RealLQ250 (DreamClear)} \\
\cmidrule(lr){2-4} \cmidrule(lr){5-8}
& PSNR$\uparrow$ & MANIQA$\uparrow$ & CLIPIQA$\uparrow$ 
& NIQE$\downarrow$ & MANIQA$\uparrow$ & MUSIQ$\uparrow$ & CLIPIQA$\uparrow$ \\
\midrule
LLaVA-13b & 21.57 & 0.4974 & 0.6184 & 3.922 & \textbf{0.4369} & 66.55 & 0.6827 \\
VisualQuality-R1 & 22.38 & 0.4898 & 0.5927 & 3.968 & 0.4273 & 65.55 & 0.6846 \\
Q-Insight & \textbf{22.44} & 0.4861 & 0.5754 & 3.965 & 0.4297 & 65.96 & 0.6820 \\
\textbf{Zoom-IQA} & 21.31 & \textbf{0.5455} & \textbf{0.6773} 
& \textbf{3.914} & 0.4365 & \textbf{66.97} & \textbf{0.6946} \\
\bottomrule
\end{tabular}
}
\vspace{-2mm}
\end{table}
Zoom-IQA attains the best CLIPIQA on both setups, the best MANIQA under SUPIR, and the best NIQE and MUSIQ under DreamClear. Moreover, supplementary DreamClear comparisons reveal superior recovery of high-frequency details.

\begin{table*}[t!]
\centering
\vspace{-2mm}
\caption{Ablation studies on each component with PLCC / SRCC metrics. Models are trained on KonIQ.}
\vspace{-4pt}
\label{tab:ablation_results}
\renewcommand{\arraystretch}{1.25}
\renewcommand{\tabcolsep}{1.0mm}
\resizebox{\textwidth}{!}{%
\begin{tabular}{@{}c|ccccc|ccccccc@{}}
\toprule
Metric &
SFT &
\begin{tabular}[c]{@{}c@{}}Score \\ Reward\end{tabular} &
\begin{tabular}[c]{@{}c@{}}Rank \\ Reward\end{tabular} &
\begin{tabular}[c]{@{}c@{}}KL-Coverage \\ Regularizer Loss\end{tabular} &
\begin{tabular}[c]{@{}c@{}}Prog. \\ Training\end{tabular} &
KonIQ & SPAQ & KADID & PIPAL & LIVE-Wild & AGIQA & CSIQ \\
\midrule

\multirow{6}{*}{PLCC}
& \checkmark &  &  &  &  & 0.836 & 0.849 & 0.632 & 0.431 & 0.806 & 0.766 & 0.688 \\
& \checkmark &  & \checkmark & \checkmark &  & 0.906 & 0.892 & 0.709 & 0.450 & 0.854 & 0.784 & 0.772 \\
& \checkmark & \checkmark &  & \checkmark &  & 0.928 & 0.896 & 0.677 & 0.379 & 0.875 & 0.807 & 0.715 \\
& \checkmark & \checkmark & \checkmark &  &  & 0.908 & 0.888 & 0.683 & 0.455 & 0.874 & 0.806 & 0.759 \\
& \checkmark & \checkmark & \checkmark & \checkmark &  & 0.932 & 0.898 & 0.665 & 0.458 & 0.881 & 0.810 & 0.791 \\
& \checkmark & \checkmark & \checkmark & \checkmark & \checkmark & 0.938 & 0.902 & 0.701 & 0.468 & 0.887 & 0.816 & 0.797 \\
\midrule

\multirow{6}{*}{SRCC}
& \checkmark &  &  &  &  & 0.806 & 0.839 & 0.621 & 0.426 & 0.762 & 0.697 & 0.645 \\
& \checkmark &  & \checkmark & \checkmark &  & 0.887 & 0.884 & 0.699 & 0.447 & 0.827 & 0.739 & 0.710 \\
& \checkmark & \checkmark &  & \checkmark &  & 0.915 & 0.892 & 0.664 & 0.392 & 0.865 & 0.745 & 0.696 \\
& \checkmark & \checkmark & \checkmark &  &  & 0.890 & 0.882 & 0.669 & 0.446 & 0.847 & 0.738 & 0.718 \\
& \checkmark & \checkmark & \checkmark & \checkmark &  & 0.918 & 0.895 & 0.652 & 0.455 & 0.865 & 0.758 & 0.749 \\
& \checkmark & \checkmark & \checkmark & \checkmark & \checkmark & 0.922 & 0.900 & 0.700 & 0.465 & 0.870 & 0.765 & 0.754 \\
\bottomrule
\end{tabular}%
}
\vspace{-2mm}
\end{table*}

\subsection{Ablation Study}

We conduct a comprehensive ablation study to evaluate the contribution of each proposed component. The results, measured by PLCC and SRCC across seven commonly used benchmarks, are presented in Table \ref{tab:ablation_results}.
Our analysis begins with the SFT model (Row 1), which serves as the baseline.

\noindent \textbf{Impact of Reward Signals.} 
The comparison between Row 2 (Rank Reward only) and Row 3 (Score Reward only) indicates the effectiveness of Rank Reward for synthetic benchmarks (e.g., KADID: 0.709 vs. 0.677; CSIQ: 0.772 vs. 0.715). Besides, the Score Reward benefits real-world benchmarks (e.g., KonIQ: 0.928 vs. 0.906; SPAQ: 0.896 vs. 0.892). Such a comparison shows that both rewards contribute to the assessment of image quality.

\noindent \textbf{Effect of KL-Coverage Regularizer Loss.}
We evaluate the impact of our KL-Coverage loss by comparing Row 4 (without KL) to Row 5 (with KL). Introducing such a regularizer brings consistent performance gains across the majority of datasets, such as on KonIQ (0.908 to 0.932), SPAQ (0.888 to 0.898), and CSIQ (0.759 to 0.791), indicating its effectiveness in preventing mode collapse and encouraging diverse reasoning.

\noindent \textbf{Effect of Progressive Training.}
Comparing Row 5 (full model without) with Row 6 (full model with), progressive training provides a consistent performance lift across all datasets. Such an improvement stems from the strategy's ability to effectively handle imbalanced quality data, \ie, conducting a superior evaluation on images at the extremes of the quality spectrum.

\subsection{Discussion}

\begin{table}[t]
  \centering
  \caption{Ablation studies on cropping strategies and training paradigms.}
  \label{tab:additional_result}
  \vspace{-3pt}
  \renewcommand{\arraystretch}{1.0}
  \renewcommand{\tabcolsep}{5.0mm}
  \makebox[\columnwidth][c]{%
    \resizebox{\columnwidth}{!}{%
      \begin{tabular}{lcccc}
        \toprule
        Setting & KonIQ & KADID & LIVE-Wild & CSIQ \\
        \midrule
        Central Crop     & 0.898/0.876 & 0.672/0.661 & 0.871/0.842 & 0.743/0.677 \\
        Single-Stage SFT & 0.834/0.812 & 0.614/0.613 & 0.801/0.775 & 0.672/0.637 \\
        Ours             & 0.938/0.922 & 0.701/0.700 & 0.887/0.870 & 0.797/0.754 \\
        \bottomrule
      \end{tabular}%
    }%
  }
  \vspace{-3pt}
\end{table}

\noindent \textbf{Necessity of the Learned Zooming Policy.}
Our learned policy consistently outperforms a simple heuristic that always zooms into the image center (``Center Crop''), while keeping the rest of the pipeline unchanged (Tab.~\ref{tab:additional_result}; e.g., +0.040 PLCC on KonIQ). This clear margin suggests that quality-relevant distortions are not necessarily centered, and that RL-based active zooming effectively helps locate degradation-critical regions.

\noindent \textbf{Saliency Grounding of Zoomed Regions.}
To evaluate whether the generated crops are semantically meaningful, we introduce a Saliency Grounding Benchmark using QAGNet~\cite{deng2024advancing}. 
We assess performance using two metrics: Saliency Density Lift (SDL)—measuring the relative saliency concentration within the crop compared to the whole image—and Tight-Coverage ($F_{0.5}$), which jointly rewards high saliency coverage and compact crop sizes. 
As shown in Tab.~\ref{tab:saliency_grounding}, the baseline Qwen2.5-VL-7b outputs excessively large crops (Area $\sim 65\%$) that far overstep the actual salient regions due to its limited grounding capabilities. In contrast, Zoom-IQA precisely localizes necessary and meaningful regions, achieving much tighter crops (Area $\sim 36\%$) alongside substantially higher SDL and $F_{0.5}$ (e.g., $0.78$ vs. $0.06$ on SPAQ). This demonstrates that our zoom-in mechanism is highly semantics-aware rather than blindly cropping.

\begin{table}[t]
\centering
\caption{Saliency Grounding Benchmark on KonIQ and SPAQ.}
\vspace{-5pt}
\label{tab:saliency_grounding}
\scriptsize
\setlength{\tabcolsep}{4.0mm} 
\renewcommand{\arraystretch}{1.0}
\resizebox{\columnwidth}{!}{%
\begin{tabular}{lcccccc}
\toprule
\multirow{2}{*}{\textbf{Method}} & \multicolumn{3}{c}{\textbf{KonIQ}} & \multicolumn{3}{c}{\textbf{SPAQ}} \\
\cmidrule(lr){2-4} \cmidrule(lr){5-7}
& Area & $F_{0.5}\uparrow$ & SDL$\uparrow$ & Area & $F_{0.5}\uparrow$ & SDL$\uparrow$ \\
\midrule
Qwen2.5-VL-7b & 0.65 & 0.19 & 1.04 & 0.66 & 0.06 & 1.01 \\
\textbf{Zoom-IQA} & \textbf{0.36} & \textbf{0.70} & \textbf{1.90} & \textbf{0.37} & \textbf{0.78} & \textbf{2.13} \\
\bottomrule
\end{tabular}
}
\vspace{-4mm}
\end{table}

\noindent \textbf{Superiority of RL over Pure Supervised Fine-Tuning.}
Fine-tuning the backbone VLM on GR-IQA using a ``Single-Stage SFT'' baseline (without the multi-stage zoom pipeline or RL) consistently underperforms our full method (Tab.~\ref{tab:additional_result}; e.g., +0.110 SRCC on KonIQ). These results indicate that while GR-IQA provides useful supervision for learning basic zooming/grounding behaviors, supervised imitation alone is insufficient to induce an effective region-selection strategy. In contrast, the RL objective enables self-guided exploration and improves the learned policy's ability to identify quality-critical regions, leading to better generalization across datasets.

\vspace{-2mm}
\section{Conclusion}
In this work, we proposed Zoom-IQA, a novel IQA framework that uses an iterative process of reasoning and zooming to focus on quality-relevant regions and generate accurate chain-of-thought reasoning. To encourage reliable reasoning, we first build the fine-grained Grounded-Rationale-IQA (GR-IQA) dataset. We further present key training strategies, including a two-stage scheme, reward designs, a KL-Coverage regularizer, and progressive resampling.
We verify the effectiveness of our designs via extensive experiments from multiple aspects, including score prediction, reasoning examination, and the downstream application, together with a thorough ablation study.
We believe our work could motivate future development in various domains, such as designing IQA data pipelines with automated data labeling, enhancing the reasoning reliability of IQA, and building more robust, interactive perceptual models.

\vspace{-2mm}
\section{Acknowledgment}
This research is supported by cash and in-kind funding from NTU S-Lab and industry partner(s). It is also supported by Singapore MOE AcRF Tier 2 (MOE-T2EP20224-0003).

\clearpage
\bibliographystyle{splncs04}
\bibliography{main}

\clearpage
\setcounter{page}{1}
\begin{center}
{\Large\bf Zoom-IQA: Image Quality Assessment with Reliable Region-Aware Reasoning}\\[2mm]
{\Large\bf Supplementary Material}
\end{center}
\vspace{3mm}

\definecolor{headergray}{gray}{0.80}  
\definecolor{sectiongray}{gray}{0.92} 
\definecolor{examplegray}{gray}{0.88} 

\section{Prompt Templates for GR-IQA Dataset Construction and Filtering}
\subsection{Data Construction Prompts}
Tab.~\ref{tab:prompt_stage1_final} and \ref{tab:prompt_stage2_final} detail the prompts used to construct our Grounded-Rationale-IQA (GR-IQA) dataset with Gemini-2.5-pro~\cite{comanici2025gemini}. The thinking rationale comprises four components:
\begin{enumerate}
    \item \textbf{Image Quality Summary:} A direct assessment of technical quality.
    \item \textbf{Directions for Improvement:} An aspirational description of the ideal image.
    \item \textbf{Issues to Avoid:} A detailed description of existing technical artifacts.
    \item \textbf{Decision \& Rationale:} A comprehensive justification covering the initial rating, crop analysis, and final decision.
\end{enumerate}
A key feature of our design is that both the \textit{Directions for Improvement} and \textit{Issues to Avoid} are \textbf{supported by regional findings}. This spatial grounding allows the reasoning path to be effectively split into positive and negative prompts, boosting performance in downstream tasks.

\subsection{Visual Reliance Filtering}
Due to the limited API access of closed-source VLMs, we cannot compute the log-probabilities for predefined text sequences (or specific candidate answers).
Consequently, existing hallucination detection methods relying on offline contrastive probability analysis~\cite{xing2025scalecap} are inapplicable.
Furthermore, prior online decoding strategies~\cite{leng2024mitigating,huo2024self} typically inject image distortions to verify consistency. While effective for semantic-level tasks—where content identity remains robust to noise—this approach is fundamentally incompatible with IQA. Since IQA aims to precisely evaluate visual degradation, introducing artificial distortion alters the target attribute itself. This makes it difficult to disentangle whether the model is responding to the original image artifacts or the injected noise.

\begin{figure*}[!t]
\centering
    \includegraphics[width=0.99\linewidth]{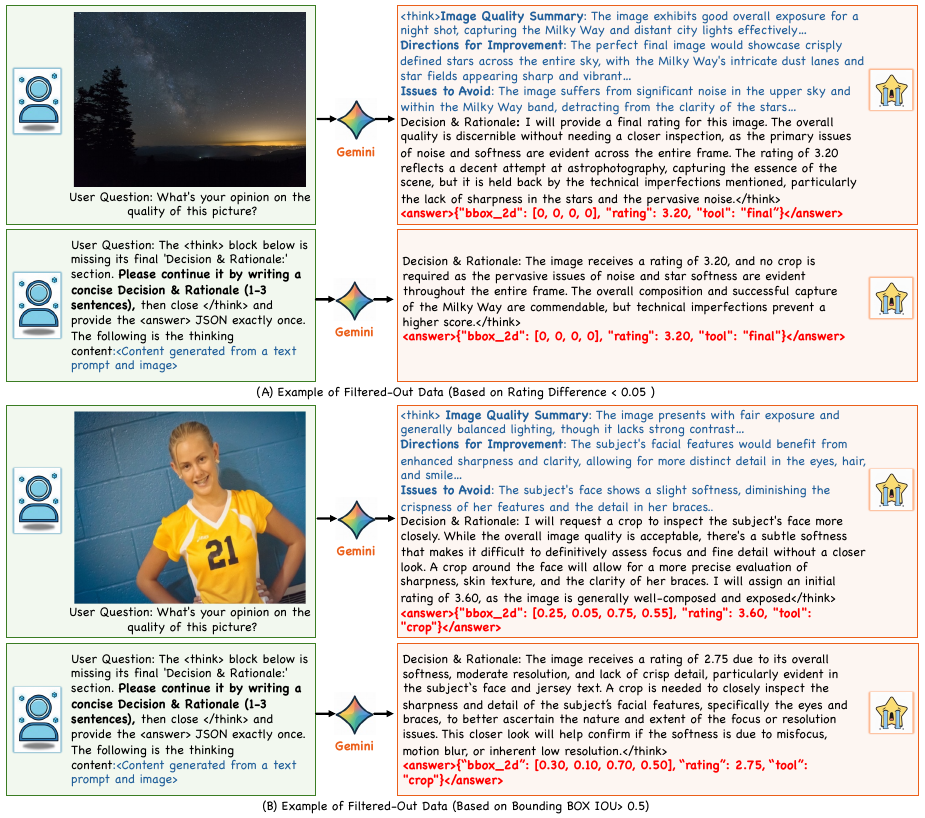}
    \caption{Example of filtered-out data in Visual Reliance Filtering.}
    \label{fig:suppl_data_1}
\centering
\end{figure*}

To address these limitations, we propose Visual Reliance Filtering (VRF). As illustrated in Fig.~\ref{fig:suppl_data_1}, VRF filters out samples where the model exhibits low visual dependency. We compare the VLM's outputs under two distinct conditions: \textbf{1)} conditioned on both the image $I$ and the textual rationale $R_A$ and \textbf{2)} conditioned only on the partial rationale $R_A$ (without visual input).
If the outputs are excessively similar, we discard the sample, as this indicates the visual input $I$ was non-essential and the response was driven primarily by language priors. In our experiments, we set the thresholds for rating difference, Bounding Box IoU, and entropy difference to 0.05, 0.5, and 0.01, respectively.

\subsection{Hint-Augmented Consistency Filtering}
Tab.~\ref{tab:data_filter_out} presents examples of data discarded by our Hint-Augmented Consistency Filtering with Qwen-2.5-32b~\cite{qwen2.5}. Specifically, we leverage the low-level hints from KonIQ~\cite{hosu2020koniq} and the quality hints from KonIQ++~\cite{su2021koniq++} as reference; this allows our method to effectively filter out generated prompts that are inconsistent with human labels.

\section{Image Quality Reasoning}
Tab.~\ref{tab:prompt_judge_final} presents the prompts used to validate the effectiveness of our reasoning chains on the KonIQ and SPAQ~\cite{fang2020perceptual} datasets. 
These datasets provide low-level image attributes (brightness, contrast, colorfulness, and sharpness for KonIQ; brightness, colorfulness, contrast, noisiness, and sharpness for SPAQ) alongside MOS scores, which ensures more precise evaluation judgments.

\subsection{Ruling Out Self-Preference Bias in Evaluation}
\noindent To avoid Gemini's ``self-preference'' in evaluation, we provide additional independent lines of evidence, as shown in Tab.~\ref{tab:integrated_results}. We add Grok-4-1-fast as a third independent evaluator and conduct a human ranking study with 15 annotators on 30 images. Both confirm the advantage of Zoom-IQA.
\begin{table}[h]
\centering
\scriptsize
\setlength{\tabcolsep}{1pt}
\renewcommand{\arraystretch}{1.2}
\renewcommand{\tabcolsep}{1.4mm}
\vspace{-2mm}
\caption{Results of LLM and Human Evaluation.}
\label{tab:integrated_results}
\begin{tabular}{@{}llccccc@{}}
\toprule
\textbf{Data} & \textbf{Method} & Accuracy $\uparrow$ & Reasonableness $\uparrow$ & Completeness $\uparrow$ & Confidence $\uparrow$ & Hum.$\downarrow$ \\
\midrule
\multirow{3}{*}{KonIQ}
& VisualQuality-R1         & 8.14 & 8.21 & 7.46 & 8.21 & 2.31 \\
& Q-Insight      & 8.11 & 8.33 & 7.19 & 7.73 & 2.58 \\
& \textbf{Ours}  & \textbf{8.49} & \textbf{8.84} & \textbf{8.37} & \textbf{8.98} & \textbf{1.11} \\
\midrule
\multirow{3}{*}{SPAQ}
& VisualQuality-R1         & 7.98 & 7.79 & 7.14 & 8.46 & 2.28 \\
& Q-Insight      & 7.84 & 7.75 & 6.88 & 7.56 & 2.65 \\
& \textbf{Ours}  & \textbf{8.62} & \textbf{8.70} & \textbf{8.37} & \textbf{8.97} & \textbf{1.07} \\
\bottomrule
\end{tabular}
\vspace{-2mm}
\end{table}

\section{More Reasoning-guided Restoration Results}
\begin{figure*}[!t]
\vspace{-20pt}
\centering
    \includegraphics[width=0.99\linewidth]{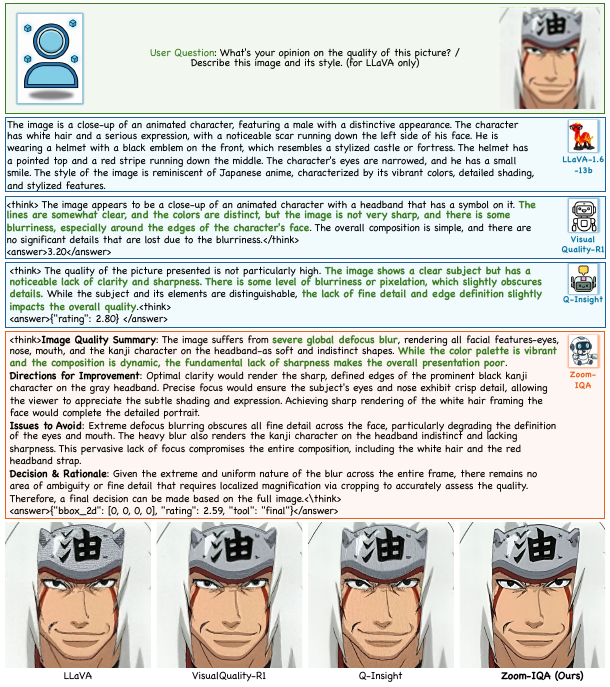}
    \vspace{-10pt}
    \caption{Qualitative comparison of Zoom-IQA against competing methods (Q-Insight~\cite{li2025q}, VisualQuality-R1~\cite{wu2025visualquality}) and the baseline (LLaVA-1.6-13b~\cite{liu2023visual}) on the image restoration task. \textbf{(Upper)} The text guidance generated by each method, with \textcolor[HTML]{3B7D23}{accurate} descriptions highlighted. \textbf{(Lower)} The corresponding restored results utilizing these text prompts. \textbf{Please zoom in for better details.}}
    \label{fig:suppl_restoration_1}
\centering
\end{figure*}

\begin{figure*}[!t]
\vspace{-20pt}
\centering
    \includegraphics[width=0.99\linewidth]{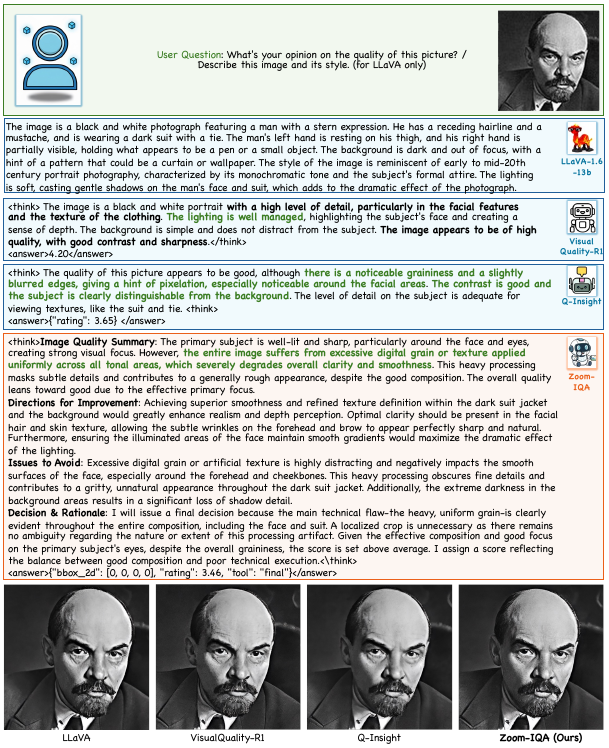}
    \vspace{-10pt}
    \caption{Qualitative comparison of Zoom-IQA against competing methods (Q-Insight~\cite{li2025q}, VisualQuality-R1~\cite{wu2025visualquality}) and the baseline (LLaVA-1.6-13b~\cite{liu2023visual}) on the image restoration task. \textbf{(Upper)} The text guidance generated by each method, with \textcolor[HTML]{3B7D23}{accurate} descriptions highlighted. \textbf{(Lower)} The corresponding restored results utilizing these text prompts. \textbf{Please zoom in for better details.}}
    \label{fig:suppl_restoration_2}
\centering
\end{figure*}

\begin{figure*}[!t]
\vspace{-20pt}
\centering
    \includegraphics[width=0.99\linewidth]{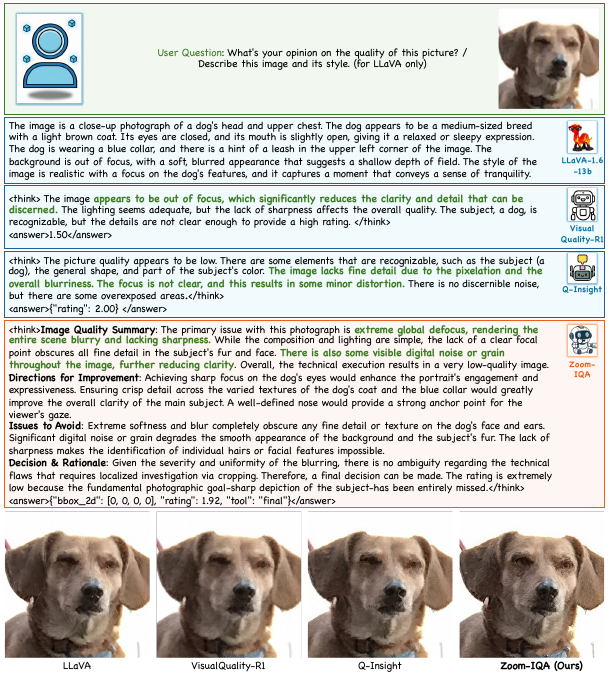}
    \vspace{-10pt}
    \caption{Qualitative comparison of Zoom-IQA against competing methods (Q-Insight~\cite{li2025q}, VisualQuality-R1~\cite{wu2025visualquality}) and the baseline (LLaVA-1.6-13b~\cite{liu2023visual}) on the image restoration task. \textbf{(Upper)} The text guidance generated by each method, with \textcolor[HTML]{3B7D23}{accurate} descriptions highlighted. \textbf{(Lower)} The corresponding restored results utilizing these text prompts. \textbf{Please zoom in for better details.}}
    \label{fig:suppl_restoration_3}
\centering
\end{figure*}

As illustrated in Figs.~\ref{fig:suppl_restoration_1}-\ref{fig:suppl_restoration_3}, restoration guided by our method's reasoning exhibits superior perceptual texture quality, particularly in complex regions such as facial features (Fig.~\ref{fig:suppl_restoration_2}) and fur (Fig.~\ref{fig:suppl_restoration_3}).

\section{More Qualitative Comparison Results}
\begin{figure*}[!t]
\vspace{-20pt}
\centering
    \includegraphics[width=0.99\linewidth]{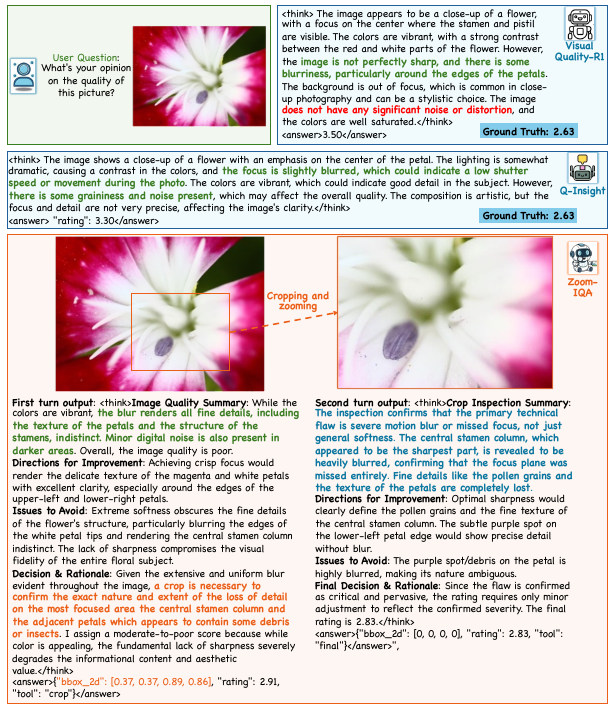}
    \vspace{-5pt}
    \caption{Qualitative comparison of Zoom-IQA with competing methods (Q-Insight~\cite{li2025q} and VisualQuality-R1~\cite{wu2025visualquality}). We highlight \textcolor[HTML]{3B7D23}{correct} descriptions and \textcolor{red}{incorrect} descriptions, in addition to the \textcolor[HTML]{E97132}{uncertainty-aware} and \textcolor[HTML]{0B76A0}{verifying} reasoning unique to our model. The \textcolor[HTML]{E97132}{bbox} indicates a large, cropped zoom requested by Zoom-IQA, \textbf{clearly showing the blurriness}. While both Q-Insight and our method correctly identify both the blurriness and noise, VisualQuality-R1 only \textbf{recognizes the blurriness and ignores the noise}. Crucially, Q-Insight provides only general distortion information, whereas our method \textbf{not only specifies the distortion types but also precisely identifies the object or region suffering from the distortion}. \textbf{Please zoom in for more details.}}
    \label{fig:suppl_comp_1}
\centering
\end{figure*}

\begin{figure*}[!t]
\vspace{-20pt}
\centering
    \includegraphics[width=0.99\linewidth]{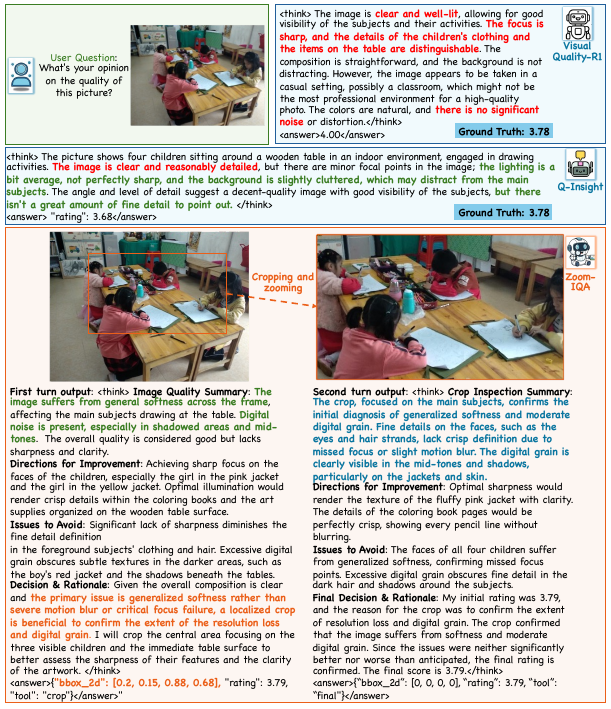}
    \vspace{-5pt}
    \caption{Qualitative comparison of Zoom-IQA with competing methods (Q-Insight~\cite{li2025q} and VisualQuality-R1~\cite{wu2025visualquality}). We highlight \textcolor[HTML]{3B7D23}{correct} descriptions and \textcolor{red}{incorrect} descriptions, along with the \textcolor[HTML]{E97132}{uncertainty-aware} and \textcolor[HTML]{0B76A0}{verifying} reasoning unique to our model. The \textcolor[HTML]{E97132}{bbox} indicates a large, cropped zoom requested by Zoom-IQA, \textbf{clearly showing the digital noise}. While Zoom-IQA accurately identifies the issues, VisualQuality-R1 \textbf{fails to detect minor focus problems and digital noise}, claiming there is no significant noise and that details are distinguishable. Furthermore, although Q-Insight notices the softness problem, it reaches \textbf{a contradictory conclusion} (asserting the image has ``reasonable detail'' but ``not a great amount of fine detail to point out'') and also \textbf{fails to recognize the digital noise}. \textbf{Please zoom in for more details.}}
    \label{fig:suppl_comp_2}
\centering
\end{figure*}

\begin{figure*}[!t]
\centering
    \includegraphics[width=0.99\linewidth]{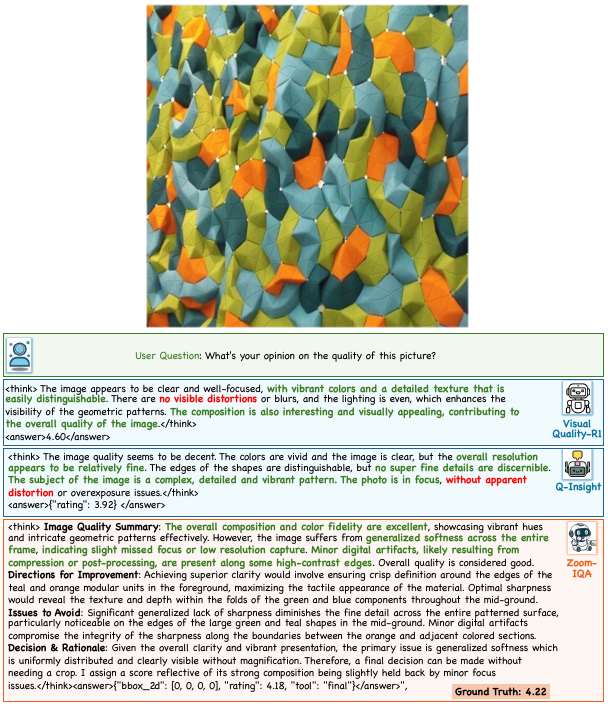}
    \caption{Qualitative comparison of Zoom-IQA with competing methods (Q-Insight~\cite{li2025q}, VisualQuality-R1~\cite{wu2025visualquality}). We highlight: \textcolor[HTML]{3B7D23}{\textbf{correct}} descriptions and \textcolor{red}{\textbf{incorrect}} descriptions. The image exhibits minor digital artifacts, which \textbf{were uniquely identified by Zoom-IQA}. \textbf{Please zoom in for more details.}}
    \label{fig:suppl_comp_3}
\centering
\end{figure*}

\begin{figure*}[!t]
\centering
    \includegraphics[width=0.99\linewidth]{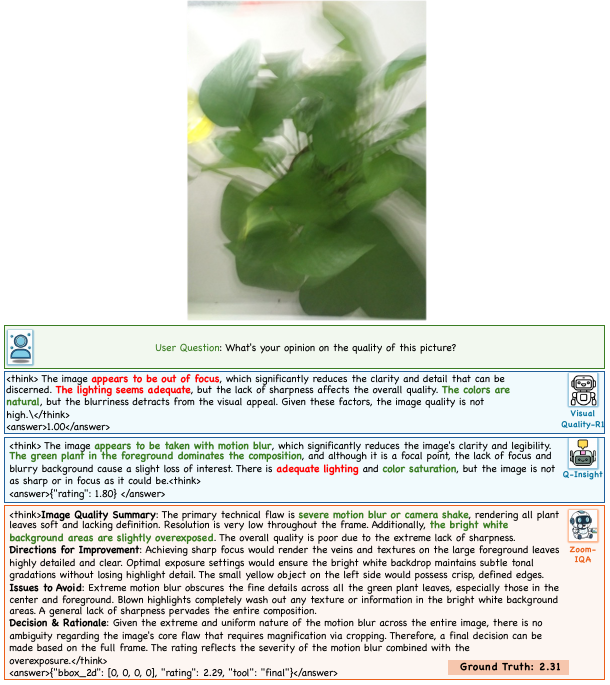}
    \caption{Qualitative comparison of Zoom-IQA with competing methods (Q-Insight~\cite{li2025q}, VisualQuality-R1~\cite{wu2025visualquality}). We highlight: \textcolor[HTML]{3B7D23}{\textbf{correct}} descriptions and \textcolor{red}{\textbf{incorrect}} descriptions. The image exhibits slight overexposure, which \textbf{was uniquely identified by Zoom-IQA}. Furthermore, VisualQuality-R1 incorrectly classified \textbf{motion blur as out-of-focus blur}. \textbf{Please zoom in for more details.}}
    \label{fig:suppl_comp_4}
\centering
\end{figure*}

We provide additional qualitative comparisons between our method and competing methods (Q-Insight~\cite{li2025q}, VisualQuality-R1~\cite{wu2025visualquality}) on real-world images, as illustrated in Fig.~\ref{fig:suppl_comp_1} through Fig.~\ref{fig:suppl_comp_4}. These results demonstrate the superiority of our reasoning mechanism, which not only identifies specific distortions but also explicitly localizes the distorted objects/regions. Furthermore, in complex scenes (\eg, Fig.~\ref{fig:suppl_comp_1} and Fig.~\ref{fig:suppl_comp_2}), our method employs interactive, region-aware reasoning: it first hypothesizes potential flaws (\textcolor[HTML]{3B7D23}{green text}), then grounds them via adaptive cropping (\textcolor[HTML]{E97132}{orange text}), and finally verifies the degradation (\textcolor[HTML]{0B76A0}{blue text}). This hypothesize-and-verify loop ensures a comprehensive assessment. Conversely, in scenes with simpler compositions (\eg, Fig.~\ref{fig:suppl_comp_3} and Fig.~\ref{fig:suppl_comp_4}), our method directly detects global distortions without performing cropping operations.

\section{Discussion}
\subsection{Robustness to Background Degradation: Handling Night-time Noise and Bokeh Artifacts}
\begin{figure*}[!t]
\centering
    \includegraphics[width=0.99\linewidth]{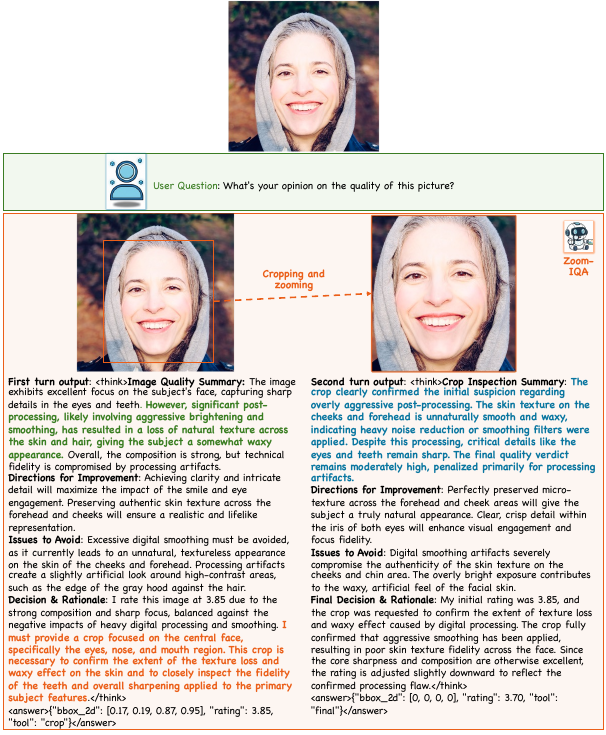}
    \caption{A real-world example with background bokeh effect.}
    \label{fig:suppl_bokeh}
\centering
\end{figure*}

\begin{figure*}[!t]
\centering
    \includegraphics[width=0.99\linewidth]{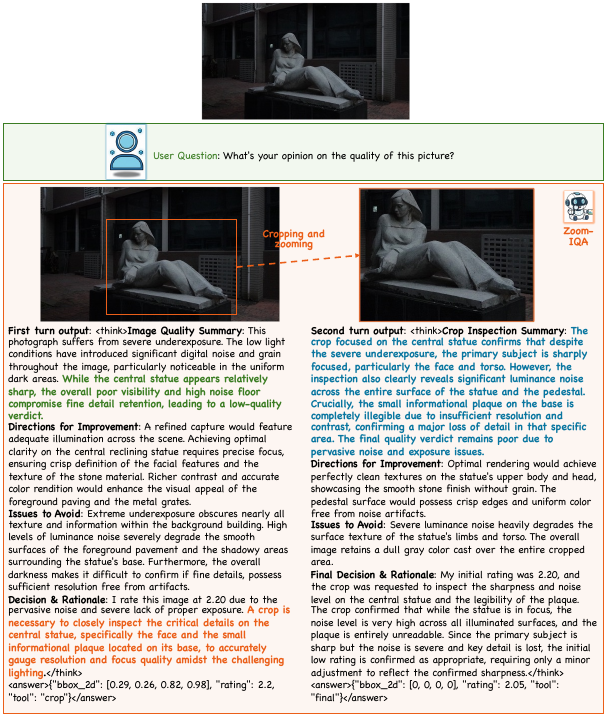}
    \caption{A real-world example with noise and low light distortion from the FoundIR~\cite{li2025foundir} test dataset.}
    \label{fig:suppl_lowlight}
\centering
\end{figure*}

A natural concern with any crop-based IQA pipeline is whether spurious background regions — such as noisy skies in night-time photography or heavily defocused bokeh from large-aperture lenses — might unduly influence the quality score. We address both concerns with representative examples.
\textbf{Large-aperture background bokeh.} A common issue arises when shallow depth-of-field blur is present in the background; penalising it would confuse artistic choice with technical deficiency. As shown in Fig.~\ref{fig:suppl_bokeh}, the model crops to the primary subject — the human face — and identifies actual artifacts like digital smoothing, while ignoring the blurred background. This ensures that background bokeh does not artificially degrade the image quality score.
\textbf{Night-time noise in background regions.} Another concern is that our model might penalise low-light images with noisy dark backgrounds, artificially lowering the score. As shown in Fig.~\ref{fig:suppl_lowlight}, the saliency-guided crop correctly focuses on the statue and pedestal, not the featureless sky. The modest score change (2.20 → 2.05) is due to subject-level noise on the illuminated surfaces, not background darkness. Non-informative dark areas are naturally excluded from the inspection window.
These cases confirm that the crop-and-zoom paradigm directs quality-sensitive inspection to regions where perceptual fidelity truly matters, suppressing irrelevant background influences on the final score.

\begin{table}[!h]
  \centering
\renewcommand{\arraystretch}{1.25}
\renewcommand{\tabcolsep}{2.0mm}
  \caption{\footnotesize Run-time comparison (sec/image).}
  \label{tab:runtime}
  \resizebox{0.75\textwidth}{!}{
  \begin{tabular}{lcccc}
    \toprule
    \textbf{Dataset} & \textbf{DeQA} & \textbf{VisualQuality-R1} & \textbf{Q-Insight} & \textbf{Ours} \\
    \midrule
    LiveW & 0.021 & 0.046 & 0.047 & 0.068 \\
    KADID & 0.028 & 0.045 & 0.041  & 0.061 \\
    \bottomrule
  \end{tabular}
  }
\vspace{-1pt}
\end{table}

\section{Inference time}
We report the inference time on the LiveW~\cite{ghadiyaram2015live} and KADID~\cite{lin2019kadid} datasets using an NVIDIA H200 NVL GPU with vLLM~\cite{kwon2023efficient}. 
As shown in Tab.~\ref{tab:runtime}, our multi-step zoom mechanism introduces only a marginal increase in latency ($\sim$0.06s versus 0.02--0.05s for the baselines). This demonstrates that the method remains computationally efficient for practical deployment while enabling superior \textbf{region-aware reasoning}.

\begin{table*}[t]
\centering
\caption{Prompt for Initial Image Quality Assessment (Stage 1). This prompt directs the model to act as an image quality expert, perform a preliminary evaluation, and decide whether a high-resolution crop is necessary to resolve uncertainties before making a final judgment.}
\label{tab:prompt_stage1_final}
\renewcommand{\arraystretch}{1.9} 
\begin{tabular}{@{}>{\bfseries}p{0.2\textwidth} p{0.75\textwidth}@{}}
\toprule
Component & Description \\ \midrule
\rowcolor{sectiongray}
Objective & To perform an initial assessment of an image's technical quality and identify a specific region of uncertainty that requires closer inspection. \\
\addlinespace[5pt]
Persona & The model is instructed to act as an \textbf{Image Quality Expert}. \\
\addlinespace[5pt]
\rowcolor{sectiongray}
Output Structure & \parbox[t]{\linewidth}{The output is a two-part structure: \\
\textbf{\texttt{<think>}}: Contains the detailed reasoning process. \\
\textbf{\texttt{<answer>}}: Contains a machine-readable JSON object with the final decision.} \\
\addlinespace[5pt]
Reasoning Sections \newline \textit{(in \texttt{<think>})} & \parbox[t]{\linewidth}{The \texttt{<think>} block must contain exactly four labeled paragraphs: \\
\textbf{1. Image Quality Summary:} A concise verdict on technical flaws. \\
\textbf{2. Directions for Improvement:} Aspirational description of a perfect image (positive framing). \\
\textbf{3. Issues to Avoid:} Description of existing technical problems (negative framing). \\
\textbf{4. Decision \& Rationale:} The initial rating, the crop/final decision, and its justification.} \\
\addlinespace[5pt]
\rowcolor{sectiongray}
Answer JSON Format \newline \textit{(in \texttt{<answer>})} & \parbox[t]{\linewidth}{The \texttt{<answer>} block contains a JSON object with the following keys: \\
\textbf{\texttt{"bbox\_2d"}:} Coordinates \texttt{[x1, y1, x2, y2]} for the crop, or \texttt{[0,0,0,0]} if no crop is needed. \\
\textbf{\texttt{"rating"}:} The initial quality score (e.g., \texttt{3.50}). \\
\textbf{\texttt{"tool"}:} Either \texttt{"crop"} to request a zoom-in, or \texttt{"final"} to conclude the assessment.} \\
\addlinespace[5pt]
Key Constraints & \parbox[t]{\linewidth}{- Reasoning must be in compact paragraphs, not bullet points. \\
- Sections 2 and 3 must maintain strictly positive and negative language, respectively. \\
- All feedback must be grounded in specific, named regions of the image.} \\ \bottomrule
\end{tabular}
\vspace{3mm}
\end{table*}

\begin{table*}[ht]
\centering
\caption{Prompt for Final Image Quality Assessment with Crop (Stage 2). This prompt is used after a crop has been generated in Stage 1. It instructs the model to synthesize information from the original image, the crop, and its own prior reasoning to produce a definitive and well-justified final quality score.}
\label{tab:prompt_stage2_final}
\renewcommand{\arraystretch}{1.9}
\begin{tabular}{@{}>{\bfseries}p{0.2\textwidth} p{0.75\textwidth}@{}}
\toprule
Component & Description \\ \midrule
\rowcolor{sectiongray}
Objective & To re-evaluate image quality using a high-resolution crop of a previously identified uncertain region and to provide a definitive, justified final rating. \\
\addlinespace[5pt]
Persona & The model continues to act as an \textbf{Image Quality Expert}. \\
\addlinespace[5pt]
\rowcolor{sectiongray}
Input Context & The model receives the original image, the crop, and its own reasoning from Stage 1. \\
\addlinespace[5pt]
Reasoning Sections \newline \textit{(in \texttt{<think>})} & \parbox[t]{\linewidth}{The \texttt{<think>} block is restructured to focus on the new evidence from the crop: \\
\textbf{1. Crop Inspection Summary:} A summary of what the crop confirmed or revealed. \\
\textbf{2. Directions for Improvement:} Aspirational goals based on details now visible within the crop. \\
\textbf{3. Issues to Avoid:} Technical problems confirmed or newly discovered within the crop. \\
\textbf{4. Final Decision \& Rationale:} Explicitly references the initial rating and explains how the crop's findings led to a rating upgrade, downgrade, or confirmation.} \\
\addlinespace[5pt]
\rowcolor{sectiongray}
Answer JSON Format \newline \textit{(in \texttt{<answer>})} & \parbox[t]{\linewidth}{The JSON output is now always final: \\
\textbf{\texttt{"bbox\_2d"}:} Always \texttt{[0, 0, 0, 0]}. \\
\textbf{\texttt{"rating"}:} The final, definitive quality score (e.g., \texttt{4.25}). \\
\textbf{\texttt{"tool"}:} Always \texttt{"final"}.} \\
\addlinespace[5pt]
In-Context Learning & \parbox[t]{\linewidth}{The prompt includes three detailed examples demonstrating how to handle different scenarios: \\
\textbf{1. Rating Downgrade:} When the crop reveals the quality is worse than suspected. \\
\textbf{2. Rating Upgrade:} When the crop reveals the quality is better than suspected. \\
\textbf{3. Rating Confirmation:} When the crop confirms the initial assessment.} \\ \bottomrule
\end{tabular}
\end{table*}

\begin{table*}[htbp]
    \centering
    \caption{Example of Filtered-Out Data in Hint-Augmented Consistency Filtering.}
    \label{tab:data_filter_out}

    \resizebox{\textwidth}{!}{%
    \begin{tabularx}{1.15\textwidth}{
        >{\raggedright\arraybackslash}p{1.75cm}
        >{\raggedright\arraybackslash}X
        >{\raggedright\arraybackslash}X
    }
        \toprule
        \textbf{Example} & \textbf{1} & \textbf{2} \\
        \midrule

        \textbf{Image} & 
        \includegraphics[width=0.90\linewidth, valign=c]{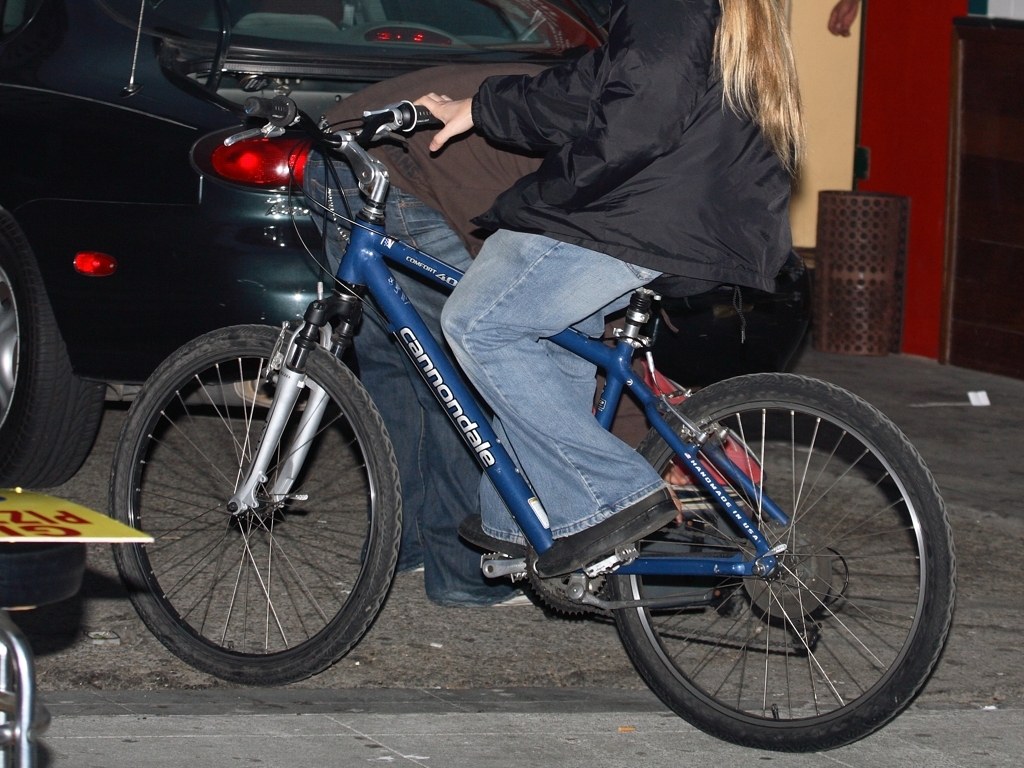} & 
        \includegraphics[width=0.90\linewidth, valign=c]{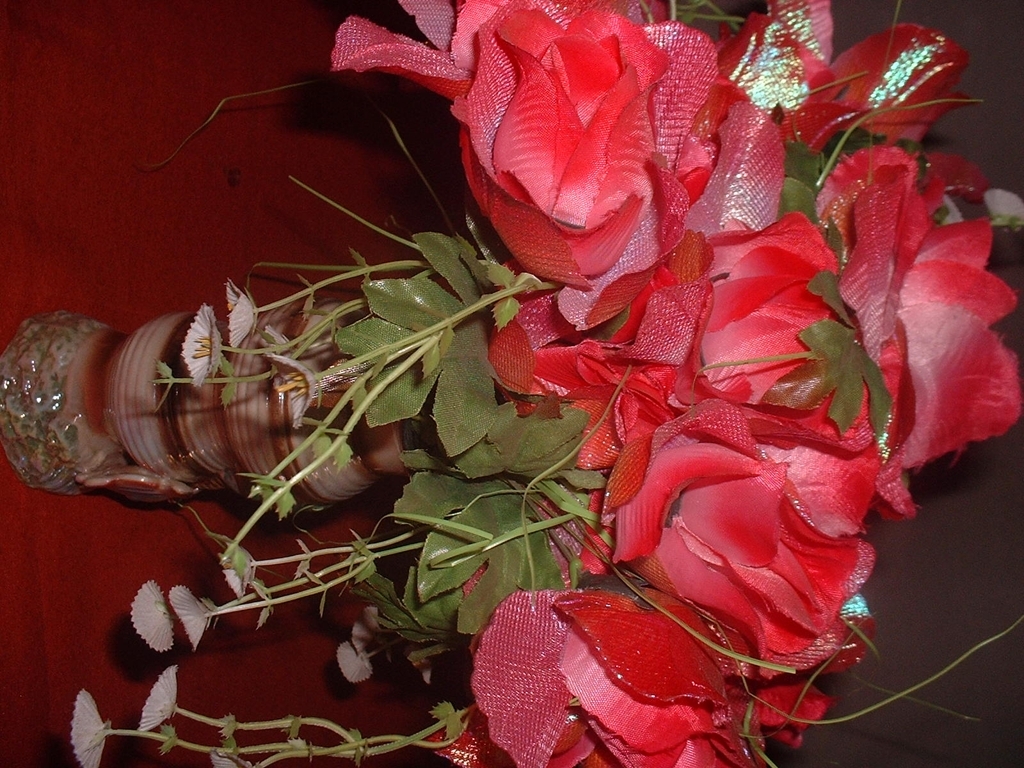} \\
        \midrule

        \textbf{Generated prompt} & 
        {\scriptsize \RaggedRight 
        
        \textbf{\textless{}think\textgreater} \par
        1) \textbf{Image Quality Summary:} The image suffers from significant digital noise, which is evident across the entire frame, especially in darker areas like the car and the person's jacket. The lighting is harsh due to direct on-camera flash, creating strong specular highlights on reflective surfaces and deep shadows, resulting in poor dynamic range. \textcolor{red}{The overall focus is soft, lacking the crispness needed to resolve fine details.} \par
        \vspace{0.3em}
        
        2) \textbf{Directions for Improvement:} A higher quality image would render the ``Cannondale'' text on the bicycle frame with perfect sharpness and legibility. The person's black jacket would appear with smooth, clean textures, and the silver fork of the bicycle would be detailed without harsh glare. The fine spokes of the wheels would be individually distinct and clear. \par
        \vspace{0.3em}

        3) \textbf{Issues to Avoid:} There is excessive digital noise across the image, particularly degrading the texture on the car's body and the person's jacket. The direct flash causes blown-out specular highlights on the bicycle's silver components and the car's taillight. A general softness affects the entire scene, preventing sharp focus on key elements like the bicycle frame and the rider's clothing. \par
        \vspace{0.3em}
        
        4) \textbf{Decision \& Rationale:} The image's technical flaws, namely the pervasive noise, harsh lighting, and soft focus, are global and clearly visible without needing a closer look. The provided hint crops confirm these issues are present in key areas of detail. Therefore, a final rating can be given without cropping. The combination of these significant problems warrants a low rating, placing the image in the ``Poor'' category. \par
        \textbf{\textless{}/think\textgreater} \par

        \vspace{0.5em}
        
        \textbf{\textless{}answer\textgreater} \par 
        \textbf{BBox 2D:} [0, 0, 0, 0] \par           
        \textbf{Rating:} 2.25 \par                    
        \textbf{Tool:} Final \par                     
        \textbf{\textless{}/answer\textgreater}
        } 
        & 
        {\scriptsize \RaggedRight 
        
        \textbf{\textless{}think\textgreater} \par
        1) \textbf{Image Quality Summary:} The image quality is poor due to a combination of technical issues. \textcolor{red}{It suffers from a soft focus across the entire frame, with no clear point of sharpness, and a very shallow depth of field.} There is significant digital noise, particularly noticeable in the darker background areas, which gives the image a grainy texture. The lighting is harsh, likely from a direct flash, creating unflattering specular highlights on the flowers and vase and resulting in an overall flat appearance. \par
        \vspace{0.3em}
        
        2) \textbf{Directions for Improvement:} A well-executed photograph would feature sharp focus on the central red roses, clearly defining the texture of their fabric petals. Diffused, soft lighting would elegantly highlight the brown and white patterns of the ribbed vase without creating harsh glare. Greater depth of field would render the delicate small white flowers with clarity, adding textural variety and depth to the floral arrangement. \par
        \vspace{0.3em}

        3) \textbf{Issues to Avoid:} The image exhibits a pervasive softness that leaves the petals of the red roses and the details of the small white flowers blurry and undefined. Harsh, blown-out highlights from a direct flash are present on the shiny, iridescent ribbons within the bouquet and across the glossy surface of the vase. \par
        \vspace{0.3em}
        
        4) \textbf{Decision \& Rationale:} I will make a final decision as the image's quality issues are immediately apparent and consistent throughout the frame. The softness, prominent noise, and harsh lighting are global problems that a crop would not help to clarify; it would only magnify these existing flaws. \par
        \textbf{\textless{}/think\textgreater} \par

        \vspace{0.5em}
        
        \textbf{\textless{}answer\textgreater} \par 
        \textbf{BBox 2D:} [0, 0, 0, 0] \par           
        \textbf{Rating:} 1.75 \par                    
        \textbf{Tool:} Final \par                     
        \textbf{\textless{}/answer\textgreater}
        }  \\
        \midrule

        \textbf{Low-level hint} & 
        {\scriptsize\RaggedRight
        MOS $\uparrow$: 3.80 \par  
        Brightness $\uparrow$: 0.2579 \par 
        Contrast $\uparrow$: 0.1604 \par 
        Colorfulness $\uparrow$: 0.3846 \par 
        Sharpness $\uparrow$: 19.23}
        & 
        {\scriptsize\RaggedRight
        MOS $\uparrow$: 3.68 \par  
        Brightness $\uparrow$: 0.2662 \par 
        Contrast $\uparrow$: 0.1779 \par 
        Colorfulness $\uparrow$: 0.0521 \par 
        Sharpness $\uparrow$: 20.07}
        \\
        \midrule

        \textbf{Quality hint} & 
        {\scriptsize\RaggedRight
        QMOS $\uparrow$: 4.01 \par  
        Artifacts $\downarrow$: 0.0000 \par 
        Blurriness $\downarrow$: 0.033}
        & 
        {\scriptsize\RaggedRight
        QMOS $\uparrow$: 3.85 \par  
        Artifacts $\downarrow$: 0.098 \par 
        Blurriness $\downarrow$: 0.066}
        \\
        \bottomrule
    \end{tabularx}%
    }
\end{table*}

\clearpage

\begin{table*}[ht]
\centering
\caption{Prompt for the ``Image Quality Reasoning'' Evaluation Strategy. This prompt configures a model to act as an expert human evaluator. It explicitly defines the priority of visual evidence over objective metrics and specifies the handling of strictly logical conclusions for both single and multi-round reasoning.}
\label{tab:prompt_judge_final}
\renewcommand{\arraystretch}{1.4}
\begin{tabular}{@{}>{\bfseries}p{0.2\textwidth} p{0.78\textwidth}@{}}
\toprule
Component & Description \\ \midrule
\rowcolor{sectiongray}
Objective & To evaluate the reasoning quality of a generative model (supporting both single-round and multi-round outputs) by scoring it against a structured, human-centric rubric. For multi-round cases, the evaluation focuses on the final conclusion and its logical evolution. \\
\addlinespace[5pt]
Persona & The model is instructed to act as an \textbf{Expert Image Quality Evaluator}, utilizing the provided image as the \textbf{primary source of truth} while treating objective metrics only as supporting technical references. \\
\addlinespace[5pt]
\rowcolor{sectiongray}
Evaluation Framework & \parbox[t]{\linewidth}{The model must score the response on a scale of [1-9] across four criteria: \\ \textbf{1. Completeness:} Does the assessment identify the \textbf{most significant perceptual qualities} a human would notice? (Reference indicators serve as a checklist). \\ \textbf{2. Accuracy:} Is the description \textbf{true to the visual evidence} first and foremost? A subjective assessment that matches human perception is prioritized over one that blindly matches metrics. \\ \textbf{3. Reasonableness:} Is the reasoning logical? Does the final conclusion \textbf{feel holistically appropriate} from a human perspective, bridging visual evidence to the assessment? \\ \textbf{4. Confidence:} Assesses the certainty of the language used (e.g., decisive declarative statements vs. hedging), regardless of the assessment's correctness.} \\
\addlinespace[5pt]
Input Context & \parbox[t]{\linewidth}{The evaluator is provided with: \\ \textbf{1. [Model Response]:} The text generated by the target model (assessing either direct reasoning or the final conclusion of a multi-round process). \\ \textbf{2. [Reference Indicators]:} Objective metrics for cross-referencing: MOS, Sharpness, Brightness, Contrast, and Colorfulness.} \\
\addlinespace[5pt]
\rowcolor{sectiongray}
Output Specification & \parbox[t]{\linewidth}{The output must be \textbf{ONLY} an XML structure containing four scores and a point-by-point justification for each: \\ \texttt{<Completeness>}[1-9]\texttt{</Completeness>} \\ \texttt{<Accuracy>}[1-9]\texttt{</Accuracy>} \\ \texttt{<Reasonableness>}[1-9]\texttt{</Reasonableness>} \\ \texttt{<Confidence>}[1-9]\texttt{</Confidence>} \\ \texttt{<think>} \\ \texttt{  - Completeness: [Brief justification]} \\ \texttt{  - ... (Justifications balancing subjective \& technical view)} \\ \texttt{</think>}} \\ \bottomrule
\end{tabular}
\end{table*}

\end{document}